\documentclass[journal]{IEEEtran}
\usepackage{cite}
\usepackage{epsfig}
\usepackage[caption=false,font=footnotesize]{subfig}
\usepackage{graphicx}
\usepackage{amssymb}
\usepackage{array}
\newcolumntype{C}[1]{>{\centering\let\newline\\\arraybackslash\hspace{0pt}}m{#1}}
\usepackage{pifont}
\usepackage[cmex10]{amsmath, nccmath}
\usepackage{mathtools}
\usepackage{upgreek}
\usepackage{graphicx}
\usepackage{algorithm,algorithmic}
\usepackage{enumitem}
\usepackage{bm}
\newtheorem{theorem}{Theorem}

\newtheorem{remark}{Remark}

\newtheorem{lemma}{Lemma}
\newtheorem{assumption}{Assumption}
\newtheorem{definition}{Definition}
\newtheorem{corollary}{Corollary}

\newcommand{\norm}[1]{\left\lVert#1\right\rVert}

\newcommand{\E}[1]{\mathbb E\left[#1\right]}
\newcommand{\Prob}[1]{\mathbb P\left[#1\right]}

\newcommand{\bs}[1]{\boldsymbol{#1}}
\newcommand{\mc}[1]{\mathcal{#1}}

\newcommand{\mr}[1]{\mathrm{#1}}

\newcommand{\thmref}[1]{Theorem~\ref{#1}}

\newcommand{\secref}[1]{Section~\ref{#1}}
\newcommand{\lemref}[1]{Lemma~\ref{#1}}

\newcommand{\corref}[1]{Corollary~\ref{#1}}
\newcommand{\appref}[1]{Appendix~\ref{#1}}

\newcommand{\figref}[1]{Fig.~\ref{#1}}

\DeclareMathOperator*{\argmax}{arg\,max}
\DeclareMathOperator*{\argmin}{arg\,min}

\usepackage{hyperref}
\usepackage{bookmark}
\usepackage{url}
\usepackage{cases}
\hypersetup{colorlinks=true, linkcolor=blue, citecolor=blue, urlcolor = blue}
\usepackage{ifthen}
\newif\ifshowtodo
\showtodotrue

\renewcommand{\IEEEproofindentspace}{1\parindent}

\newcommand{\VersionLength}{long}
\providecommand{\ver}{\ifthenelse{\equal{\VersionLength}{long}}}
\allowdisplaybreaks
\begin{document}
\title{A General Framework for Clustering and Distribution Matching 
with Bandit Feedback}
\author{Recep Can Yavas,~\IEEEmembership{Member,~IEEE}, Yuqi Huang, Vincent Y. F. Tan,~\IEEEmembership{Senior Member,~IEEE}, \\and Jonathan Scarlett,~\IEEEmembership{Member,~IEEE}
\thanks{Manuscript received October 23, 2024; revised January 1, 2025; accepted January 8, 2025.}
\thanks{When the manuscript was submitted, R. C. Yavas was with the Descartes Program, CNRS@CREATE, 138602, Singapore. He is now with the Department of Computer Science, National University of Singapore, 119077, Singapore (e-mail: ryavas@nus.edu.sg). Y. Huang is with the Department of Information Systems and Analytics, School of Computing, National University of Singapore, 119077, Singapore (e-mail: e0727232@u.nus.edu). V. Y. F. Tan is with the Department of Mathematics and Department of Electrical and Computer Engineering,
National University of Singapore, 119077, Singapore (e-mail: vtan@nus.edu.sg). J. Scarlett is with the Department of Computer Science, Department of Mathematics, and Institute of Data Science, National University of Singapore, 119077, Singapore (e-mail: scarlett@comp.nus.edu.sg).}}
\IEEEoverridecommandlockouts
\maketitle 
\begin{abstract}
We develop a general framework for clustering and distribution matching problems with bandit feedback. We consider a $K$-armed bandit model where some subset of $K$ arms is partitioned into $M$ groups. 
Within each group, the random variable associated to each arm follows the same distribution on a finite alphabet. At each time step, the decision maker pulls an arm and observes its outcome from the random variable associated to that arm. Subsequent arm pulls depend on the history of arm pulls and their outcomes. The decision maker has no knowledge of the distributions of the arms or the underlying partitions. The task is to devise an online algorithm to learn the underlying partition of arms with the least number of arm pulls on average and with an error probability not exceeding a pre-determined value~$\delta$. Several existing problems fall under our general framework, including finding $M$ pairs of arms, odd arm identification, and $N$-ary clustering of $K$ arms belong to our general framework. We derive a non-asymptotic lower bound on the average number of arm pulls for any online algorithm with an error probability not exceeding $\delta$. Furthermore, we develop a computationally-efficient online algorithm based on the Track-and-Stop method and Frank--Wolfe algorithm, and show that the average number of arm pulls of our algorithm asymptotically matches that of the lower bound. Our refined analysis also uncovers a novel bound on the speed at which the average number of arm pulls of our algorithm converges to the fundamental limit as $\delta$ vanishes. 
\end{abstract}
\section{Introduction} \label{sec:intro}
Dating back to Thompson \cite{thompson} and Chernoff \cite{chernoff1959}, multi-armed bandits (MABs) provide a useful framework for sequential design of experiments on $K$ unknown distributions. At each time step, a learner chooses an arm based on the history of the experiments and receives a reward from the chosen arm. Two of the most common objectives in MABs are \emph{regret minimization}, where the goal is to maximize the expected value of the total received reward at a time $T$ (see \cite{bubeckbook} for a survey), and \emph{pure exploration} (PE), where the goal is to answer a specific question on the $K$ unknown distributions. Examples of PE problems include best arm identification (BAI) \cite{garivier2016}, where the goal is to identify the arm with the largest mean; $\epsilon$-good arm identification \cite{mason2020epsilon}, where the goal is to identify all arms whose means are within $\epsilon$ distance to the largest mean; odd arm identification \cite{vaidhiyan2018, karthik2019}, where the goal is to identify the arm that follows a distribution different from the rest; and clustering \cite{yang2024clustering, thuot2024active}, where the goal is to learn a hitherto unknown partitioning of arms. Two settings are considered for PE problems: \emph{fixed confidence} \cite{evendar2002} and \emph{fixed budget} \cite{bubeck2009Budget}. In the fixed-confidence setting, the decision is made at a random stopping time $\tau$ with the goal of minimizing the expected value of $\tau$ while ensuring that the error probability does not exceed a predetermined value $\delta$. In the fixed-budget setting, the decision is made at a fixed time $T$, and the goal is to develop an online algorithm with an error probability as small as possible. As discussed in \cite{prabhu2022}, PE problems can generally be viewed as sequential multi-hypothesis testing with bandit feedback since each potential answer to the question of interest defines a hypothesis. Active sequential hypothesis testing, studied in~\cite{naghshvarActive}, in which the decision maker can choose one of $K$ actions at each time step to eventually declare one of $M$ hypotheses as the true one, is also closely related to PE problems.   

Clustering problems have drawn significant attention in the data analysis and machine learning literature due to their wide range of applications including bioinformatics, pattern recognition, and commercial decision-making. In commercial decision-making, accurately clustering customers into groups based on their user profiles is crucial for the success of the online recommendation systems, where the customer feedback is collected in an online manner. In the case of an unknown type of virus (e.g., COVID-19), when an accurate laboratory analysis of the virus variants is not yet available, clustering the noisy measurements from the infectious patients into specific virus variants can help healthcare professionals combat the virus. For more motivating examples of clustering with or without bandit feedback, see \cite{yang2024clustering} and \cite{thuot2024active} and the references therein.  

\subsection{Problem Setting}
In this work, we study a PE problem that broadly involves clustering of arms and/or finding the arms whose distributions are \emph{matched}. We assume that arms follow unknown distributions that are supported  on a common finite alphabet. We consider the fixed-confidence setting from a sequential multi-hypothesis testing perspective. Specifically, each hypothesis corresponds to a particular partitioning of $K$ arms into $M + 1$ groups, where the first $M$ groups are called \emph{clusters} and the remaining group is called the \emph{unconstrained group}. We assume that each cluster contains at least two arms, and the distributions of the arms in a cluster are identical. 
The arms in the unconstrained group may or may not share the same distribution as one another. Moreover, the unconstrained group may be empty depending on the specific structure of the problem. Fixing the number of clusters to the same value for all hypotheses is not critical to our results; however, doing so simplifies the presentation. 
We study a general framework for clustering, meaning that the hypotheses (i.e., partitions) included in the problem depend on the structure of the specific task of our interest. A motivating scenario where this setup may be of interest is as follows. At each time, the decision maker queries one of the $K$ users to choose an item from a common finite set of items. The user then chooses an item randomly according to a distribution supported on that finite set. This distribution determines the user's profile. The users' item choices are revealed to the decision maker at each time. The decision maker aims to cluster the user profiles as quickly as possible. In this setting, the users are the arms, and the items are the outcomes in MABs.

\begin{figure}
\center
\includegraphics[width=1\linewidth]{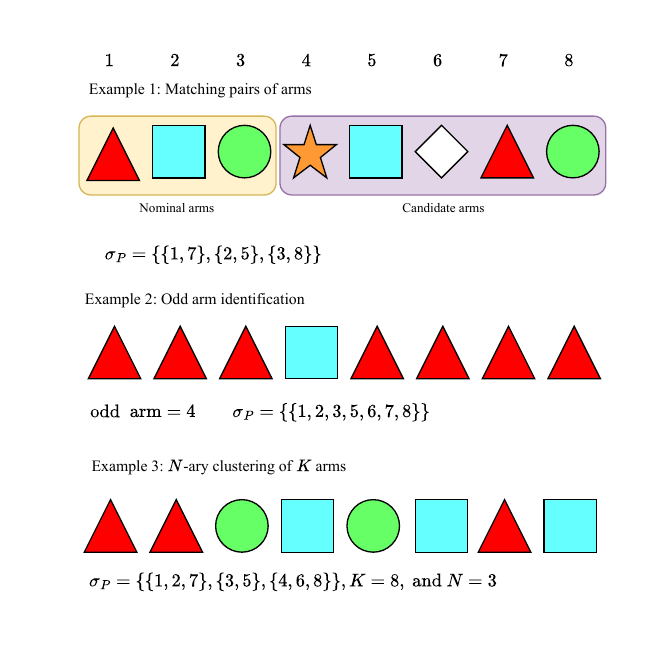}
\caption{The examples of matching pairs, odd arm identification, and $N$-ary clustering of $K$ arms are illustrated. Each shape indicates a unique distribution. Arms that are demonstrated by the same shape are in the same cluster. The number of arms for each example is $K = 8$. For Examples 1, the number of clusters is $M = 3$. In Example 1, the decision maker knows that arms 1, 2, and 3 are the nominal arms that must appear in $M = 3$ clusters. In Example 2, the decision maker knows that exactly one arm has a different distribution than the others. In Example 3, the decision maker knows that there are $N = 3$ individual groups but does not know about the size or the content of each group.}
\label{fig:examples}
\end{figure}

To make our framework more concrete, we present three special cases that are encompassed by our framework. These three examples are illustrated in Fig.~\ref{fig:examples}.
\begin{enumerate}
    \item \emph{Matching pairs with two groups of arms:} In this problem, we consider $M$ nominal arms and $K-M$ candidate arms with $K \geq 2M$. For each nominal arm $i \in [M]$, there exists exactly one candidate arm $j \in \{M + 1, \dots, K\}$ such that arm $i$ and arm $j$ follow the same distribution. The decision maker is aware of which arms are nominal/candidate but the arm distributions and the $M$ matches are unknown.  
    The goal is to identify all $M$ pairs of arms with the matching distributions. This problem is the bandit feedback adaptation of the statistical sequence matching problem considered in \cite{unnikrishnan, zhou2024, zhou2024arxiv} where data is collected offline. The offline version in Zhou et al. \cite{zhou2024, zhou2024arxiv}, which generalizes the one in Unnikrishnan \cite{unnikrishnan}, is motivated by the need for accurate user targeting in advertisement recommendation systems. 
    In \cite{zhou2024, zhou2024arxiv}, there are two databases, consisting of $M_1$ and $M_2$ training sequences. The lengths of each sequence in databases 1 and 2 are $N$ and $n$, respectively. The goal is to design a test that outputs the $M \leq \min\{M_1, M_2\}$ pairs (one from database 1 and one from database 2) that are generated from the same distribution. In \cite{zhou2024, zhou2024arxiv}, the authors consider a scenario in which the test has an option to reject any of the hypotheses, and they study the trade-off between the false reject and mismatch error probabilities as $n$ and $N$ approach infinity, where the ratio of sequence lengths $\frac{N}{n}$ is fixed to some $\alpha > 0$. They consider both the cases where $M$ is known and unknown and derive the fundamental limits in the small and large deviations regimes.  

    In our problem, data is collected in an online and sequential manner, and the decision is made at a random stopping time $\tau$ as soon as the decision maker has sufficient confidence in her estimate on the unknown true hypothesis. Hence, the number of pulls for arm $i$ up to time $\tau$ is a random variable, which makes the problem and analysis substantially different from the one in \cite{zhou2024, zhou2024arxiv}. Unlike in \cite{zhou2024, zhou2024arxiv}, we do not consider the rejection option, and we fix the number of matches to $M$.  
    \item \emph{Generalized odd arm identification:} In the standard odd arm identification problem introduced by Vaidhiyan and Sundaresan \cite{vaidhiyan2018}, the goal is to identify the anomalous arm (referred to as the ``odd'' arm) that follows a distribution that is different from the rest of the $K-1$ arms. In our framework, this problem has $M = 1$ cluster of $K-1$ arms, and the unconstrained group contains only a single arm, the odd arm. In \cite{vaidhiyan2018}, the arms are assumed to follow an independent and identically distributed Poisson process; in \cite{karthik2019}, Karthik and Sundaresan consider Markov arms. Our formulation generalizes this problem to scenarios with $M > 1$ clusters and/or multiple odd arms. 
    \item \emph{$N$-ary clustering of $K$ arms:} In \cite{yang2024clustering}, Yang et al.\ introduce an online clustering problem where $K$ arms are partitioned into $N \geq 2$ groups\footnote{The number of groups $N$ in Example 3 is not always equal to the number of clusters in the above definition. Therefore, we use a different letter than~$M$.}, and each group in the partitioning can have any size greater than or equal to~1. Since a cluster has at least two arms according to our definition, the number of clusters in each hypothesis, denoted by $M_{\sigma}$ becomes the difference between $N$ and the number of groups of size 1 in that hypothesis, instead of $N$. The groups of size 1 are collected together in the \emph{unconstrained group}. This problem is unstructured in the sense that any partitioning of $K$ arms into $N$ groups of any sizes defines a valid hypothesis. 
    This is in contrast with scenarios such as Example 1, where
    the labeling of arms matters since the decision maker knows which arms are the nominal arms and that the nominal arms must be present in the respective ``clusters''. 
    
    Yang et al.\ \cite{yang2024clustering} consider a setting in which each arm follows a $d$-dimensional Gaussian distribution with an unknown mean and known identity covariance matrix. The high-dimensional scenario where the number of groups $N$ and the dimension $d$ can be as large as $\log \frac{1}{\delta}$ is studied in \cite{thuot2024active}.
    For a fixed $N$, the number of hypotheses in this problem, equal to the Stirling number of the second kind \cite{yang2024clustering}, grows asymptotically as $N^K / N!$. To circumvent the exponentially growing number of hypotheses, using the fact that any partitioning of $K$ arms into $N$ groups defines a valid hypothesis, Yang et al.\ simplify the inner infimum of the fundamental limit (see \eqref{eq:Tstargeneral}, below) to a tractable finite minimization problem. This, together with  the Gaussianity of the arms, then aids designing and analyzing a version of the classical $K$-means algorithm to decide which arm to sample next. 
    They also propose a novel stopping rule that is based on a different statistic than the natural generalized log-likelihood ratio (GLLR) used in \cite{garivier2016}.

    In this work, our focus is on $N$-ary clustering of $K$ arms that follow distributions on a {\em finite alphabet}. Since the Kullback--Leibler (KL) divergence between two distributions on a common finite alphabet is, in general, not symmetric in its arguments, the $K$-means algorithm used in \cite{yang2024clustering}, which heavily relies on the sum of squares objective, does not directly apply to our scenario. Therefore, our theoretical result complements the result in \cite{yang2024clustering} for scenarios with arm distributions on a finite alphabet. 
    Although our sequential hypothesis testing approach does not simplify the inner infimum of the fundamental limit in \eqref{eq:Tstargeneral}, our unified algorithm is applicable  to a wide range of clustering problems where a hypothesis indicates which arms have identical distributions, and it does not rely on the symmetries that can arise from the hypotheses or the arm distributions. 
    
\end{enumerate}

\subsection{The Lower Bound and Track-and-Stop Algorithms}
For general PE problems in the fixed-confidence setting, the state-of-the-art asymptotically optimal (as $\delta$ approaches 0) algorithm utilizes a lower bound on $\mathbb{E}[\tau]$. This lower bound is first proved and used in the algorithm proposed in \cite{garivier2016} for BAI; it generalizes to more general PE problems as well \cite{prabhu2022, wang2021fast}. Let $P = (P_1, \dots, P_K)$ be a problem instance, where $P_i$ denotes the distribution of arm $i$. For a general PE problem, the lower bound yields (see \cite[Prop.~2]{wang2021fast})
\begin{align}
    \E{\tau} \geq \frac{d(\delta \| 1- \delta)}{T^*(P)} = \frac{\log \frac{1}{\delta}}{T^*(P)} + \Theta(1), \label{eq:lowergeneral}
\end{align}
where
     \begin{align}
         T^*(P)
         &= \sup_{w \in \Sigma_{K}} \inf_{P' \in \mr{Alt}(P)} \sum_{i \in [K]} w_i D(P_i \| P_i'). \label{eq:Tstargeneral}
     \end{align}
Here, $\Sigma_K$ denotes the $K$-dimensional simplex set, $\mr{Alt}(P)$ denotes the set of problem instances whose corresponding true hypothesis is different than that of the instance $P$, and $d(\cdot \| \cdot)$ denotes the KL divergence between two Bernoulli distributions. The quantity $\frac{1}{T^*(P)}$ is the hardness parameter that determines how long it takes to identify the underlying partition associated with $P$. The optimizer $P'$ of the infimum problem in \eqref{eq:Tstargeneral} corresponds to the ``most confusing'' alternative instance relative to $P$. The allocation vector $w^*(P)$ that is the maximizer in \eqref{eq:Tstargeneral} corresponds to the ideal fraction of arm pulls for the $K$ arms of any (asymptotically) optimal algorithm. By the law of large numbers, given that the algorithm guarantees that each arm is pulled sufficiently often, the empirical problem instance at time $t$, $\hat{P}(t) \triangleq (\hat{P}_1(t), \dots, \hat{P}_K(t))$, approaches $P$ almost surely as $t$ grows. Then, if the optimal allocation vector $w^*(P)$ is a continuous function of $P$, then we argue that $w^*(\hat{P}(t))$ is close to $w^*(P)$ as well. We then \emph{track} the empirically optimal allocation $w^*(\hat{P}(t))$ at each time $t$ in the sense that the empirical distribution of arm pulls at time $t$ is close to $w^*(\hat{P}(t))$. We ensure that each arm is sampled sufficiently often by adding a \emph{forced exploration} component that samples the arms uniformly at random at regular steps.

With a suitably chosen stopping rule, the algorithm with the above principle achieves an average stopping time bounded as
\begin{align}
    \E{\tau} \leq \frac{\log \frac{1}{\delta}}{T^*(P)} (1 + o(1)), \label{eq:uppergeneral}
\end{align}
showing together with \eqref{eq:lowergeneral} that $\frac{\log \frac{1}{\delta}}{T^*(P)}$ is the asymptotically optimal average stopping time. This type of algorithm is called a Track-and-Stop (\textsf{TaS}) algorithm and is developed in \cite{garivier2016}  to derive the fundamental limit for BAI with arm distributions from a single-parameter exponential family. The scaling of the ``second-order term'' $o(1)$ in \eqref{eq:uppergeneral} with respect to a decaying $\delta$ has not been derived in \cite{garivier2016, prabhu2022, wang2021fast}. Degenne et al.~\cite{degenne2019} derive a non-asymptotic bound on $\E{\tau}$ for BAI where arm distributions belong to a one-parameter exponential family. Differently than \textsf{TaS}-type algorithms, Degenne et al.'s algorithm treats the maximin problem in \eqref{eq:Tstargeneral} as an unknown game. Another non-asymptotic bound is derived in~\cite{barrier2022} for BAI where arm distributions are Gaussian with a fixed and known variance. 

Under the continuous selection assumption of the optimal allocation $w^*(P)$ with respect to $P$, Prabhu et al.~\cite{prabhu2022} develop a \textsf{TaS}-type algorithm that proves \eqref{eq:uppergeneral} not only for BAI but for general PE problems where the arm distributions are from a vector exponential family. They show that BAI and odd arm identification (the second example above) satisfy the continuous selection assumption. Despite this result, the primary drawback of Prabhu et al.'s algorithm is that the maximin problem in \eqref{eq:Tstargeneral}, which is used as an oracle, is assumed to be exactly and efficiently solved for the PE problem of interest. Unfortunately, efficiently solving \eqref{eq:Tstargeneral} can be a challenging task in itself. The inner infimum problem in \eqref{eq:Tstargeneral} turns out to be non-convex for some PE problems including BAI where the ``best'' arm is defined as the arm with the smallest conditional value-at-risk \cite{agrawal2021}. The outer supremum in \eqref{eq:Tstargeneral} is always a convex program (specifically, the maximization of a concave function over a convex set) but it might become difficult to solve depending on the structure of the problem. For example, for the standard BAI problem \cite{garivier2016}, the supremum can be solved via a simple line search strategy if the arms are from a single-parameter exponential family. However, in the case of heavy-tailed distributions, solving the inner infimum in \eqref{eq:Tstargeneral} is computationally demanding, which makes the line search strategy computationally expensive. Therefore, for the case of heavy-tailed distributions, Agrawal et al.~\cite{agrawalHeavy} propose an algorithm that solves the maximin problem in \eqref{eq:Tstargeneral}  periodically instead of at each time step. 

To deal with the PE problems for which the exact solution to \eqref{eq:Tstargeneral} is computationally demanding, Wang et al.~\cite{wang2021fast} devise a Frank--Wolfe-type Sampling (\textsf{FWS}) algorithm that is based on a modification of the Frank--Wolfe algorithm \cite{frankwolfe}. The Frank--Wolfe algorithm solves smooth convex (or concave) programs by linearizing the objective function and slowly moving towards the optimizer of the linear function in each iteration. A modification of the vanilla Frank--Wolfe algorithm is needed since the objective function of the outer supremum in \eqref{eq:Tstargeneral} is a non-smooth concave function of $w$. Specifically, the objective function is a point-wise minimum of finitely many concave functions.
Under some mild assumptions, Wang et al.\ prove that their \textsf{FWS} algorithm achieves an average number of arm pulls that is asymptotically optimal as $\delta$ vanishes. Although our framework of clustering problems satisfies \cite[Assumptions~1--3]{wang2021fast}, the vanilla \textsf{FWS} algorithm is not applicable to our clustering problem since it does not account for the scenario where the empirical problem instance $\hat{P}(t)$ does not belong to the set of instances included in the problem. 

\subsection{Contributions}
Our contributions are summarized as follows.
\begin{enumerate}
    \item We develop a general framework for online clustering and distribution matching in PE problems, where the arms follow a distribution on a common finite alphabet. Our framework allows us to study the identification of matching pairs, odd arm, and $N$-ary clusters as a single unified problem. 
    \item We develop an online algorithm, Track-and-Stop Strategy based on Frank--Wolfe Algorithm \textsf{(TaS-FW)}, which is asymptotically optimal as the target error probability $\delta$ approaches zero. Our algorithm \textsf{TaS-FW} adapts the \textsf{FWS} algorithm from \cite{wang2021fast} to our framework of clustering problems. \textsf{TaS-FW} is efficient in the sense that it only requires solving a single linear program at each time step. 
    One important difference with \cite{garivier2016, wang2021fast} is that while the forced exploration component in \cite{garivier2016, wang2021fast} ensures that each arm is pulled at least $\Omega(\sqrt{t})$ times up to time $t$, \textsf{TaS-FW} ensures that each arm is pulled at least $\Omega(\sqrt{t} \log t)$ times. This difference occurs because the empirical problem instance at time $t$, $\hat{P}(t)$, with high probability, does not correspond to any of the hypotheses in the clustering problem. 
    \item In \thmref{thm:lower}, we derive a non-asymptotic lower bound on the average number of arm pulls of any online algorithm with an error probability not exceeding $\delta$, and show that it is lower bounded by
    \begin{align}
        \frac{\log\frac{1}{\delta}}{T^*(P)} + \Theta(1).
    \end{align}
    
    In \thmref{thm:upper}, we derive a refined bound on the average number of arm pulls of \textsf{TaS-FW} and show that it is upper bounded by 
    \begin{align}
           \frac{\log\frac{1}{\delta}}{T^*(P)} \left( 1 + O\left( \left(\log \frac{1}{\delta} \right)^{-1/4} \sqrt{\log \log \frac{1}{\delta}} \right) \right) \label{eq:secondorder}
    \end{align}
      as $\delta\to0^+$.  
Our bound in \eqref{eq:secondorder} involves a novel ``second-order'' term (remainder term in $\log \frac{1}{\delta}$) that upper bounds the speed of convergence to the fundamental limit $\frac{\log\frac{1}{\delta}}{T^*(P)}$.  
Combining \thmref{thm:upper} with \thmref{thm:lower}, we prove that \textsf{TaS-FW} has an asymptotically optimal (in the leading/first order) average number of arm pulls for any problem instance in our clustering framework. To derive our results, we build on the results in \cite{wang2021fast} that analyze the convergence properties of the Frank--Wolfe algorithm suitably modified to be amenable to  non-smooth objective functions.
    \item We conduct experiments for each of the three examples above. The empirical results support the theoretical result in \thmref{thm:upper}. Although our theoretical upper bound in \thmref{thm:upper} does not apply to continuous distributions such as the Gaussian distribution, in \secref{sec:Gaussian}, we empirically compare the performances of \textsf{TaS-FW} and Yang et al.'s \textsf{BOC} algorithm \cite{yang2024clustering} for the Gaussian arm distributions. We observe that the average number of arm pulls for both algorithms is close to the theoretical lower bound.
\end{enumerate}

The paper is organized as follows. \secref{sec:notation} introduces the notation and formally defines our unified problem setting. \secref{sec:lower} presents the lower bound on the average stopping time of any $\delta$-correct online algorithm. In \secref{sec:TAS}, we present our algorithm \textsf{TaS-FW} and an asymptotic upper bound (\thmref{thm:upper}) on its average stopping time. \secref{sec:experiments} demonstrates some simulation results that support our theoretical results. \secref{sec:proof} presents the proof of the upper bound in \thmref{thm:upper}. \secref{sec:disc} discusses our algorithm and compares it with various PE algorithms, and \secref{sec:conclusion} concludes the paper.

\section{Problem Formulation} \label{sec:notation}
\subsection{Notation}
For $n \in \mathbb{N}$, we denote $[n] \triangleq \{1, \dots, n\}$ and the length-$n$ vector $x^n \triangleq (x_1, \dots, x_n)$. Given a sequence $(x_i)_{i \in \mc{I}}$, for any $\mc{A} \subseteq \mc{I}$, we denote $x_{\mc{A}} \triangleq (x_i \colon i \in \mc{A})$. For $n_1 \leq n_2 \in \mathbb{N}$, we denote $[n_1 \colon n_2] \triangleq \{n_1, \dots, n_2\}$. 
The $K$ dimensional vector with a 1 in the $k$-th coordinate and zeros in the other coordinates is denoted by $e_k$, i.e., $\{e_k\}_{k \in [K]}$ forms the standard orthogonal basis. All-zero and all-one vectors of dimension $K$ are denoted by $\bs{0}$ and~$\bs{1}$, respectively.

The set of distributions on an alphabet $\mc{X}$ is denoted by $\mc{P}(\mc{X})$. For $P$ and $Q$ from a common alphabet $\mc{X}$, the KL divergence (or the relative entropy) between $P$ and $Q$ is denoted by 
\begin{align}
    D(P \| Q) &\triangleq \sum_{a \in \mc{X}} P(a) \log \frac{P(a)}{Q(a)}.
\end{align}
The binary KL divergence is denoted by
\begin{align}
    d(p \| q) \triangleq  p \log \frac{p}{q} + (1-p) \log \frac{1-p}{1-q}.
\end{align}
The entropy of a discrete distribution $P$ is denoted by 
\begin{align}
    H(P) = \sum_{a \in \mc{X}} P(a) \log \frac{1}{P(a)}.
\end{align}

A collection of distributions indexed by $\mc{K}$ is denoted by $P_{\mc{K}} = (P_k \colon k \in \mc{K})$. For $P = (P_1, \dots, P_K)$ and $Q = (Q_1, \dots, Q_K) \in \mc{P}^{K}(\mc{X})$, we denote the $\ell_{\infty}$-norm between collection of distributions by $\norm{P - Q}_{\infty} \triangleq \max_{k \in [K]} \norm{P_k - Q_k}_{\infty}$. 

The probability simplex in $\mathbb{R}^K$ is denoted by $\Sigma_{K} \triangleq \{(w_1, \dots, w_K) \in \mathbb{R}^{K} \colon \sum_{i = 1}^K w_i = 1, w_i \geq 0 \,\forall i \in [K]\}$. For any $\eta \in [0, 1/K]$, we denote $\Sigma_K^{\eta} = \{w \in \Sigma_K \colon w_i \geq \eta \, \forall \, i \in [K]\}$. The interior and the closure of a set $\mc{A}$ are denoted by $\mr{int}(\mc{A})$ and $\mr{cl}(\mc{A})$, respectively.


The empirical distribution (or \emph{type}) of a sequence $x^n \in \mathcal{X}^n$ is defined as
\begin{align}
    \hat{P}_{x^n}(a) \triangleq \frac{1}{n} \sum_{i = 1}^n 1\{x_i = a\}, \quad a \in \mc{X}. \label{eq:type}
\end{align}
The set of types of length $n$ on alphabet $\mc{X}$ is denoted by $\mc{P}_n(\mc{X}) \triangleq \{P \in \mc{P}(\mc{X}) \colon n P(a) \in \mathbb{Z} \,\, \forall \, a \in \mc{X}\}$. The set of all sequences of length $n$ of type $Q$ is denoted by $\mc{T}_{Q}^n$.
We employ the standard $o(\cdot)$, $O(\cdot)$, $\Omega(\cdot)$, and $\Theta(\cdot)$ notations for asymptotic relationships of functions. These asymptotic notations will always refer to the limit as $\delta$ approaches zero.

\subsection{Problem Statement}
We now formally state the problem. We consider an MAB model with $K$ arms. The outcome of each arm follows a distribution $P_i$ from a finite alphabet $\mc{X}$. We assume that $p_{\min} \triangleq \min_{i \in [K]} \min_{a \in \mc{X}} P_i(a)  > 0$. The distributions of the arms, $P = (P_1, \dots, P_K)$, and $p_{\min}$ are unknown to the decision maker. We call $P$ a \emph{problem instance}. 
We consider a sequential multiple hypothesis test to decide which hypothesis about the instance $P$ is correct.

A hypothesis $\sigma$ consists of a partition of a non-empty \emph{subset} of the arm set $[K]$, where each subset in the partition has size at least two and indicates that the distributions of the arms in that subset are identical. Unless stated otherwise (see, e.g., Example 3 below), for each hypothesis $\sigma$, the number of such subsets is a pre-specified value $M$. 
Let the partition corresponding to the hypothesis $\sigma$ be denoted~by
\begin{align}
    \sigma \triangleq \{\mc{A}_1^{\sigma}, \dots, \mc{A}_M^{\sigma}\}, \label{eq:sigmapart}
\end{align}
where $\{\mc{A}_{m}^{\sigma}\}_{m \in [M]}$ are mutually disjoint and $ \cup_{m \in [M]} \mc{A}_m^{\sigma} \subseteq [K]$. Under $\sigma$, $P_{i_m} = P_{j_m}$ for all $i_m, j_m \in \mc{A}_m^{\sigma}$, and $m \in [M]$. We call each $\mc{A}_{m}^{\sigma}$, $m \in [M]$, a \emph{cluster}. 

We denote the \emph{unconstrained group} by 
\begin{align}
    \mc{A}_{M+1}^{\sigma} \triangleq [K] \setminus \bigcup_{m \in [M]} \mc{A}_m^{\sigma},
\end{align}
which means that any two arms in $\mc{A}_{M + 1}^{\sigma}$ may or may not follow the same distribution, i.e., the hypothesis $\sigma$ does not restrict the distributions of the arms in $\mc{A}_{M+1}^{\sigma}$ to be equal. The unconstrained group can be empty depending on the structure of the problem. Since $|\mc{A}_m^{\sigma}| \geq 2$ for $m \in [M]$, the unconstrained group satisfies $0 \leq |\mc{A}_{M + 1}^{\sigma}| \leq K - 2 M$, which implies that $K \geq 2 M$.

We assume that any two arms from two distinct sets from $\mc{A}_{[M+1]}^{\sigma} = \{\mc{A}_1^{\sigma}, \dots, \mc{A}_{M + 1}^{\sigma}\}$ have distinct distributions, i.e., for all $m_1 \neq m_2 \in [M + 1]$, $i_{m_1} \in \mc{A}_{m_1}^{\sigma}$, $j_{m_2} \in \mc{A}_{m_2}^{\sigma}$, we have $P_{i_{m_1}} \neq P_{j_{m_2}}$. 

The set of all problem instances that conform to hypothesis $\sigma$ is denoted by
\begin{align}
    \Lambda_{\sigma} &\triangleq \big(P \in \mc{P}^K(\mc{X}) \colon P_{i_m} = P_{j_m} \,\, \forall \, m \in [M], i_m, j_m \in \mc{A}_{m}^{\sigma}, \notag \\
    &\quad \text{ and } P_{i_{m_1}} \neq P_{j_{m_2}} \,\, \forall \, m_1 \neq m_2 \in [M + 1],\notag \\
    &\quad i_{m_1} \in \mc{A}_{m_1}^{\sigma}, j_{m_2} \in \mc{A}_{m_2}^{\sigma}\big). \label{eq:Lsigma}
\end{align}
The set $\Lambda_{\sigma}$ is convex.
A general clustering problem, denoted by $\mc{C}$, is defined as a collection of hypotheses $\sigma$, each corresponding to some problem instance set $\Lambda_{\sigma}$. The set of all problem instances for a given $\mc{C}$ is denoted by 
\begin{align}
    \Lambda \triangleq \bigcup_{\sigma \in \mc{C}} \Lambda_{\sigma}.
\end{align}

A hypothesis $\sigma$ is said to \emph{dominate} another hypothesis $\sigma'$ if every subset in the partitioning of $\sigma$ in \eqref{eq:sigmapart}, $\mc{A}_{[M]}^{\sigma}$, is a subset of some subset in the partitioning of $\sigma'$, $\mc{A}_{[M]}^{\sigma'}$. To give an example, consider two hypotheses $\sigma_1 = \{\{1, 2\}, \{4, 5\}\}$ and $\sigma_2 = \{\{1, 2, 3\}, \{4, 5\}\}$ with $M = 2$ and $K = 5$. The equality relations implied by $\sigma_1$ ($P_1 = P_2$ and $P_4 = P_5$) are contained in the equality relations implied by $\sigma_2$ ($P_1 = P_2$, $P_1 = P_3$, and $P_4 = P_5$). Hence, $\sigma_1$ dominates $\sigma_2$, and it holds that $\Lambda_{\sigma_2} \subset \mr{cl}(\Lambda_{\sigma_1})$. This means that
any problem instance that belongs to $\sigma_2$ is ``surrounded by'' problem instances that belong to $\sigma_1$. 
The following assumption is used to avoid such cases and to guarantee the distinguishability of any two hypotheses. 
\begin{assumption} \label{as:finer}
    For a given clustering problem $\mc{C}$, 
    \begin{enumerate}
        \item there exists no hypothesis pair $(\sigma, \sigma') \in \mc{C}^2$ for which $\sigma \neq \sigma'$ and $\sigma$ dominates~$\sigma'$,
        \item for each problem instance $P \in \Lambda$, there exists a unique hypothesis $\sigma \in \mc{C}$ such that $P \in \Lambda_{\sigma}$. 
    \end{enumerate}
\end{assumption}

We denote the unique hypothesis associated with $P$ by $\sigma_P$. Assumption~\ref{as:finer} ensures that each $P \in \Lambda$ is sufficiently away (in any distance metric) from all distributions that belong to some alternative hypothesis~$\sigma' \neq \sigma_P$. In other words, $\Lambda_{\sigma}$ is open in $\Lambda$. In the remainder of the paper, we focus only on the clustering problems for which Assumption~\ref{as:finer} holds. 

Given a clustering problem $\mc{C}$, the decision maker's goal is to identify the true hypothesis $\sigma_P$ corresponding to the unknown problem instance $P \in \Lambda$ as quickly as possible using an online algorithm while maintaining a confidence requirement as described below.

Let $A_t \in [K]$ be the decision maker's choice of arm at time $t \in \mathbb{N}$. The outcome at time $t$, denoted by $X_{t, A_t}$, is drawn from $P_{A_t}$.  Here, $X_{t, A_t}$ is independent of $(A_s, X_{s, A_s})_{s \neq t}$ given $A_t$. Let $\mc{F}_t$ be the sigma-algebra generated by $(A_s, X_{s, A_s})_{s \in [t]}$.

\begin{definition}
    An online algorithm $\pi$ is defined by
\begin{itemize}
    \item a sampling rule $(A_t)_{t \in \mathbb{N}}$, where $A_t$ is $\mc{F}_{t-1}$-measurable,
    \item a stopping time $\tau$ with respect to the filtration $\{\mc{F}_t\}_{t \in \mathbb{N}}$,
    \item a recommendation rule $\hat{\sigma}(\tau) \in \mc{C}$, which is $\mc{F}_{\tau}$-measurable. 
    \end{itemize}

    An algorithm $\pi$ for a problem $\mc{C}$ is said to be $\delta$-correct if $\Prob{\tau < \infty} = 1$ and $\Prob{\hat{\sigma}(\tau) \neq \sigma_P} \leq \delta$ for all problem instances $P$ for a given confidence value $\delta \in (0, 1)$.

\end{definition}

Our goal is to develop a $\delta$-correct online algorithm $\pi$ with the minimum average number of arm pulls $\E{\tau}$ possible.

In the remainder of the section, we detail the three example problems introduced in \secref{sec:intro} above. In \appref{sec:proofAs}, we show that all three examples satisfy Assumption~\ref{as:finer}.

\subsubsection{Example 1 (Matching pairs from two groups of arms)} In this problem, we consider $M$ nominal arms indexed by $[M]$ and $K - M$ candidate arms indexed by $[M + 1 \colon K]$. Let $P_i$ denote the distribution of the $i$-th arm. Each nominal arm $i \in [M]$ is matched with exactly one candidate arm $j \in [M + 1 \colon K]$. This means that there exists a one-to-one mapping $\sigma \colon [M] \to \mc{B}$ such that $\mc{B} \subseteq [M+1 \colon K]$, $|\mc{B}| = M$, and $P_i = P_{\sigma(i)}$ for all $i \in [M]$. Some candidate arms remain unmatched if $K > 2M$. With some abuse of notation, the partitioning in \eqref{eq:sigmapart} is expressed as
\begin{align}
    \sigma = \{\{1, \sigma(1)\}, \{2, \sigma(2)\}, \dots, \{M, \sigma(M)\}\}. 
\end{align}
The total number of hypotheses is $|\mc{C}| = \frac{(K-M)!}{(K - 2M)!}$, which approximately scales as $K^M$. 


\subsubsection{Example 2 (Generalization of the odd arm identification problem)} In the odd arm identification problem, $K - 1$ out of $K$ arms have identical distributions, $P_1$, and the remaining ``odd'' arm has a different distribution $P_2 \neq P_1$. The goal is to identify the index of the odd arm. The odd arm identification problem is recovered by our formulation above as follows. The number of clusters is $M = 1$, and each hypothesis $\sigma$ satisfies $|\mc{A}_1^{\sigma}| = K - 1$ and $|\mc{A}_2^{\sigma}| = 1$. The total number of hypotheses is $|\mc{C}| = K$. In our formulation, $M$ can be greater than 1, and there can be multiple odd arms that belong to none of the clusters, i.e., $|\mc{A}_{M + 1}^{\sigma}| > 1$ is allowed. Therefore, our formulation strictly generalizes the odd arm identification problem.


\subsubsection{Example 3 (\texorpdfstring{$M$}{N}-ary clustering of \texorpdfstring{$K$}{K} arms)} In this problem, $K$ arms are partitioned into $N \geq 2$ groups where within each group, the arms follow the same distribution. The size of each group can be any integer greater than or equal to~1. Any hypothesis $\sigma$ that satisfies this property is included in $\mc{C}$. The goal is to identify the unknown partitioning of the arms. For $N = 2$, the total number of hypotheses is $|\mc{C}| = 2^{K-1} - 1$. 

Since the size of each cluster is at least 2 according to our definition but the group size can be as small as 1, the number of clusters $M_{\sigma} = N -  |\mc{A}_{M_{\sigma}+1}^{\sigma}|$ depends on hypothesis $\sigma$, where $\mc{A}_{M_{\sigma}+1}^{\sigma}$ denotes the unconstrained group for $\sigma$, which collects the groups of size 1.

 \section{A Converse (Lower) Bound} \label{sec:lower}

Let $P \in \Lambda$ be a problem instance. The set of alternatives to the instance $P$ is defined as the set of instances whose associated hypothesis is different than $\sigma_P$, i.e.,
 \begin{align}
     \mr{Alt}(P) \triangleq \bigcup_{\sigma' \neq \sigma_P} \Lambda_{\sigma'}. 
 \end{align}

We introduce a quantity that plays a critical role in presenting our results. 
For a set $\mc{A} \subseteq [K]$, distributions $P_{\mc{A}}$ on the same alphabet $\mc{P}(\mc{X})$, and constants $w_{\mc{A}} \in [0, 1]^{|\mc{A}|}$, we define the function
 \begin{align}
     &G(P_{\mc{A}}, w_{\mc{A}}) \triangleq \begin{cases} 0 &\text{if } w_i = 0, \,\, \forall i \in \mc{A} \\ \sum_{i \in \mc{A}} w_i D(P_i \| W) &\text{otherwise } \end{cases}, \label{eq:GA}
 \end{align}
 where
 \begin{align}
     W \triangleq \frac{\sum_{i \in \mc{A}} w_i P_i}{\sum_{i \in \mc{A}} w_i} \in \mc{P}(\mc{X}). 
 \end{align}

We consider the following hypothesis test to give an intuition about the use of the function $G$ in general clustering problems. Let $(X_{i}^{n_i})_{i \in [B]}$ be $B \geq 2$ collection of sequences of lengths $n_1, \dots, n_B$ from a finite alphabet $\mc{X}$. Let $N = \sum_{i = 1}^B n_i$, $X^N = (X_1^{n_1}, \dots, X_M^{n_B})$, and
\begin{align}
    H_0 &\colon X^N \sim P^N \text{ for some } P \in \mc{P}(\mc{X}) \label{eq:hypoH0}\\
    H_1 &\colon X_i^{n_i} \sim P_i^{n_i}, i \in [B] \text{ for some } P_{[B]} \in \mc{P}^B(\mc{X}), \label{eq:hypoH1}
\end{align}
that is, the hypothesis $H_0$ considers that there exists a common distribution $P$ for all sequences in the collection, and the hypothesis $H_1$ considers that no such common distribution exists. 
The following result relates the function $G$ with a GLLR between the hypotheses $H_1$ and $H_0$.
 \begin{lemma} \label{lem:GLRG}
     Consider $X^N$ and the hypotheses $H_0$ and $H_1$ as defined in \eqref{eq:hypoH0}--\eqref{eq:hypoH1}. Let $w_i = \frac{n_i}{N}$ for $i \in [B]$. Denote $\hat{P}_{[B]} = (\hat{P}_{X_i^{n_i}})_{i \in [B]}$. Then,
     \begin{align}
          G(\hat{P}_{[B]}, w_{[B]}) = \frac{1}{N} \log \frac{\max \limits_{P_{[B]} \in \mc{P}^B(\mc{X})} \prod_{i = 1}^B P_i^{n_i}(X_i^{n_i})}{\max \limits_{P \in \mc{P}(\mc{X})} P^{N}(X^N)}. \label{eq:GLRT}
     \end{align}
 \end{lemma}
 \begin{IEEEproof}
    We note that the expression \eqref{eq:GLRT} for $B = 2$ is shown in \cite[Lemma~2]{haghifam}. Here, we prove the general statement.

     Recall that given a sequence $x^n$, the maximum likelihood estimator is the empirical distribution $\hat{P}_{x^n}$. Therefore, the maximum in the numerator in \eqref{eq:GLRT} is achieved at $P_{[B]} = \hat{P}_{[B]}$. Similarly, the maximum in the denominator is achieved at $P = \frac{\sum_{i \in [B]} w_i \hat{P}_i}{\sum_{i \in [B]} w_i}$. After some algebra, we get
     \begin{align}
         \log \max \limits_{P_{[B]} \in \mc{P}^M(\mc{X})} \prod_{i \in [B]} P_i^{n_i}(X_i^{n_i}) &= - \sum_{i \in [B]} n_i H(\hat{P}_i) \\
         \log {\max \limits_{P \in \mc{P}(\mc{X})} P^{N}(X^N)} &= - N G(\hat{P}_{[B]}, w_{[B]}) \notag \\
         &\quad - \sum_{i \in [B]} n_i H(\hat{P}_i),
     \end{align}
     which proves the claim.
 \end{IEEEproof}
 
 For $|\mc{A}| = 2$, the function $G(P_{[2]}, w_{[2]})$ is a scaled version of the generalized Jensen--Shannon (GJS) divergence defined in \cite{haghifam}. More precisely, $G(P_{[2]}, w_{[2]}) = w_2 \mr{GJS} \big(P_1, P_2, \frac{w_1}{w_2} \big)$. Therefore, the function $G$ further generalizes the GJS divergence. It is clear that $G(P_{[B]}, w_{[B]})$ is continuously differentiable in both $w_{[B]}$ and $P_{[B]}$.

 The following result relates the function $G$ with an optimization problem that appears in our analysis.
 \begin{lemma} \label{lem:inf}
 Fix $w_{[B]} \in [0, 1]^B$ having at least one nonzero entry, and fix $P_{[B]} \in \mc{P}^B(\mc{X})$. Then,
 \begin{align}
     \inf_{Q \in \mc{P}(\mc{X})} \sum_{i \in [B]} w_i D(P_i \| Q) = G(P_{[B]}, w_{[B]}),
 \end{align}
 and $Q^* = \frac{\sum_{i \in [B]} w_i P_i}{\sum_{i \in [B]} w_i}$ uniquely achieves the infimum.
 \end{lemma}
 \begin{IEEEproof}
     Consider any $Q \in \mc{P}(\mc{X})$. We have
     \begin{align}
         &\sum_{i \in [B]} w_i D(P_i \| Q) \notag \\
         &= \sum_{i \in [B]} \sum_{a \in \mc{X}} w_i P_i(a) \left(\log \frac{P_i(a)}{Q^*(a)} + \log \frac{Q^*(a)}{Q(a)} \right) \\
         &= \left(\sum_{i = 1}^B w_i D(P_i \| Q^*) \right) + \left(\sum_{i = 1}^B w_i \right) D(Q^* \| Q) \label{eq:Dequality}\\
         &\geq G(P_{[B]}, w_{[B]}),
     \end{align}
     where \eqref{eq:Dequality} uses the definition of $Q^*$, and the inequality follows from the non-negativity of the KL divergence and $\sum_{i \in [B]} w_i > 0$. Since $D(Q^* \| Q) = 0$ if and only if $Q^* = Q$, the claim is proved.
 \end{IEEEproof}

The following quantities are used to express our main results.
For every $P \in \mc{P}^{K}(\mc{X})$, $\sigma \in \mc{C}$, and an allocation vector $w_{[K]} = (w_1, \dots, w_K) \in \Sigma_{K}$, we define
 \begin{align}
     g^{\sigma}_{P}(w) &\triangleq \sum_{m = 1}^M G(P_{\mc{A}_m^\sigma}, w_{\mc{A}_m^{\sigma}}) \label{eq:gsV}\\
     G^{\sigma}_P(w) &\triangleq \min_{\sigma' \in \mc{C} \setminus \{\sigma\}} g^{\sigma'}_{P}(w) \label{eq:GsV} \\
     T(P, \sigma) &\triangleq \max_{w \in \Sigma_{K}} G_P^{\sigma}(w). \label{eq:TPsigma}
 \end{align}
\begin{remark} \label{rem1}
    The function $G(P_{\mc{A}_m^\sigma}, w_{\mc{A}_m^{\sigma}})$ is smooth by its definition in \eqref{eq:GA}, and is concave in $w$ since by \lemref{lem:inf}, it is the minimum of concave functions in $w$. Therefore, $g_P^{\sigma}(w)$ is also smooth and concave in $w$.
The function $G^{\sigma}_P(w)$ is also concave in $w$ since it is the minimum of concave functions but is non-smooth since it is the minimum of \emph{finite} number of functions. At points where two of those functions are equal, the function $G^{\sigma}_P(w)$ is not differentiable in~$w$.
\end{remark}

\begin{remark}
    From \lemref{lem:GLRG}, the function $g^{\sigma}_{P}(w)$ measures the GLLR between the hypothesis $H_1$ in \eqref{eq:hypoH1}, which involves no clusters, and $\sigma$, under the problem instance $P$ and allocation $w \in \Sigma_K$. By \eqref{eq:GA}, $g^{\sigma}_{P}(w) \geq 0$, and the equality is satisfied if and only if $\sigma$ is the hypothesis that corresponds to the instance $P$, i.e., $\sigma = \sigma_P$. The smaller $g_P^{\sigma}(w)$ is, the more the instance $P$ agrees with the hypothesis $\sigma$. Note that $g^{\sigma}_{P}(w)$ is a valid quantity even when $P$ does not belong to any hypothesis in~$\mc{C}$. 

    If $\sigma \in \mc{C}$ is the minimizer of $g_P^{\sigma}(w)$, which occurs when $\sigma =  \sigma_P$, then the function $G_P^{\sigma}(w)$ measures the GLLR between $\sigma$ and its most confusing alternative hypothesis under the problem instance $P$ and allocation $w$. For such $P$ and $\sigma$, the function $T(P, \sigma)$ yields the maximum (over all possible allocations of $K$ arms) GLLR between $\sigma$ and its most confusing alternative hypothesis.

    Our algorithm, described in \secref{sec:TAS}, below, uses $g_P^{\sigma}(w)$ as the score associated with the hypothesis $\sigma$ (the lower score is favored) and $G_P^{\sigma}(w)$ as the statistics to decide when to stop the algorithm and make a decision.  
\end{remark}
 
The following result, which gives a non-asymptotic lower bound on the average number of arm pulls for any $\delta$-correct online algorithm, follows steps similar to those in \cite[Th.1]{garivier2016} and uses \lemref{lem:inf}. The lower bound in \thmref{thm:lower} is not restricted to discrete arm distributions; it applies to continuous distributions such as the Gaussian distribution as well.
 \begin{theorem} \label{thm:lower}
     Let $\mc{C}$ be a problem that Assumption~\ref{as:finer} holds. For any $\delta$-correct algorithm $\pi$ with $\delta \in (0, \frac{1}{2})$ and any problem instance $P \in \Lambda$, 
     \begin{align}
         \E{\tau} \geq \frac{d(\delta \| 1- \delta)}{T^*(P)} \geq \frac{1}{T^*(P)} \log \frac{1}{2.4 \delta}, \label{eq:lowercluster}
     \end{align}
     where
     \begin{align}
         T^*(P)
         &= \sup_{w \in \Sigma_{K}} \inf_{P' \in \mr{Alt}(P)} \sum_{i \in [K]} w_i D(P_i \| P_i') \label{eq:TstarP}\\
         &= \sup_{w \in \Sigma_{K}} \min_{\sigma' \neq \sigma_P} \inf_{P' \in \Lambda_{\sigma'}} \sum_{i \in [K]} w_i  D(P_i \| P'_i) \label{eq:3opt} \\
         &= T(P, \sigma_P). \label{eq:Tmax}
     \end{align}
 \end{theorem}
 \begin{IEEEproof}
     The proof of \eqref{eq:TstarP} follows from  \cite[Th.~1]{garivier2016}, which applies to arbitrary PE problem in MABs. We provide the details here for completeness. We will see that the reduction in \eqref{eq:Tmax} follows due to a property of clustering problems. 

     Consider any $\delta$-correct algorithm $\pi$ and any instance $P$. Let $P' \in \mr{Alt}(P)$. We denote the probability measure under $P$ and $P'$ by $P[ \cdot]$ and $P'[ \cdot]$, respectively. Similarly, $\mathbb{E}_P$ and $\mathbb{E}_{P'}$ denote the expectation under the measures $P$ and $P'$. Let $\tau$ be a stopping time as defined in the problem formulation. Denote the number of pulls for arm $i \in [K]$ up to time $\tau$ by $N_i(\tau)$. By $\delta$-correctness, $P[\tau < \infty] = 1$ and $P'[\tau < \infty] = 1$. From the well-known general lower bound in \cite[Lemma~1]{kaufmann2016complexity}, for any $\mc{F}_{\tau}$-measurable event $\mc{E}$, we have
     \begin{align}
         \sum_{i \in [K]} \mathbb{E}_P[N_i(\tau)] D(P_i \| P'_i) \geq d(P[\mc{E}] \,\| \, P'[\mc{E}]),
     \end{align}
     which follows from a change of measure argument, Wald's identity, and the data processing inequality on the KL divergence. 

     We specify the event $\mc{E} = \{ \hat{\sigma}(\tau) = \sigma_{P'}\}$. Then the $\delta$-correctness of $\pi$ gives us
     \begin{align}
         d(P[\mc{E}] \, \| \, P'[\mc{E}]) \geq d(\delta \,\| \, 1 - \delta). 
     \end{align}
     Since $P'$ is an arbitrary alternative, we have
     \begin{align}
         d(\delta \,\| \, 1 - \delta) &\leq  \mathbb{E}_P[\tau] \inf_{P' \in \mr{Alt}(P)} \sum_{i \in [K]} \frac{\mathbb{E}_P[N_i(\tau)]}{\mathbb{E}_P[\tau]} D(P_i \| P'_i) \\
         &\leq  \mathbb{E}_P[\tau] \sup_{w \in \Sigma_K} \inf_{P' \in \mr{Alt}(P)} \sum_{i \in [K]} w_i  D(P_i \| P'_i ),
     \end{align}
     which proves \eqref{eq:TstarP}. Using $\mr{Alt}(P) = \cup_{\sigma' \neq \sigma_P} \Lambda_{\sigma'}$, we get \eqref{eq:3opt} as well.

    Using the definition in \eqref{eq:Lsigma}, we can write the objective function of the supremum in \eqref{eq:3opt} as
    \begin{align}
        &\min_{\sigma' \neq \sigma_P} \inf_{P' \in \Lambda_{\sigma'}} \sum_{i \in [K]} w_i  D(P_i \| P'_i ) \notag \\
        &= \min_{\sigma' \neq \sigma_P} \inf_{\substack{P', Q_1, \dots, Q_M \colon \\ P'_{\mc{A}_m^{\sigma'}} = (Q_m, \dots, Q_m) \\ \forall \, m \in [M]}} \sum_{m \in [M]} \sum_{i_m \in \mc{A}_m^{\sigma}} w_{i_m} D(P_{i_m} \| Q_{m})  \label{eq:separable} \\
        &=  \min_{\sigma' \neq \sigma_P} \sum_{m = 1}^M \inf_{Q_m} \sum_{i_m \in \mc{A}_{m}^{\sigma'}} w_{i_m} D(P_{i_m} \| Q_m) \label{eq:sep_used} \\
        &= \min_{\sigma' \neq \sigma_P} \sum_{m = 1}^M G(P_{\mc{A}_m^{\sigma'}}, w_{\mc{A}_{m}^{\sigma'}}), \label{eq:lem_opt_used}
    \end{align}
    where \eqref{eq:sep_used} follows since the optimization problem in \eqref{eq:separable} is separable for each equality relation $m \in [M]$, and \eqref{eq:lem_opt_used} follows from \lemref{lem:inf}. The last step in \eqref{eq:lem_opt_used} proves \eqref{eq:Tmax}.
 \end{IEEEproof}



\section{A Track-and-Stop Strategy Based on Frank--Wolfe Algorithm (\textsf{TaS-FW}) \\ and an Achievability (Upper) Bound} \label{sec:TAS}
In this section, we propose an algorithm (\textsf{TaS-FW}) that asymptotically achieves the lower bound in \thmref{thm:lower} as the error probability vanishes. Our algorithm is based on the \textsf{FWS} algorithm developed in \cite{wang2021fast}, which is a Track-and-Stop-type algorithm that approximately solves the oracle in \eqref{eq:TstarP} via Frank--Wolfe iterations. At the end of the section, we present our main result, which is an asymptotic achievability bound associated with the performance of \textsf{TaS-FW}.

Let $N_i(t)$ be the number of arm pulls up to time $t$ for arm $i \in [K]$, i.e., 
\begin{align}
    N_i(t) &\triangleq \sum_{s = 1}^t 1\{A_s = i\},
\end{align}
and denote the empirical allocation vector at time $t$ by
\begin{align}
    w(t) \triangleq (w_1, \dots, w_K) = \left( \frac{N_1(t)}{t}, \dots, \frac{N_K(t)}{t} \right). \label{eq:wt}
\end{align}
For brevity, we write $N(t) = (N_1(t), \dots, N_{K}(t))$ as well. 

With some abuse of notation, we denote the empirical distribution for arm $i$ by
\begin{align}
    \hat{P}_i(t) &\triangleq \hat{P}_{\tilde{X}_i^{N_i(t)}}, \label{eq:hatPi}
\end{align}
where $\tilde{X_i}^{N_i(t)} = (X_{n, A_n} \colon A_n = i, n \leq t)$ is the outcome vector for arm $i$ up to time $t$. The empirical problem instance at time $t$ is denoted by 
\begin{align}
    \hat{P}(t) = (\hat{P}_1(t), \dots, \hat{P}_{K}(t)). \label{eq:hatPdef}
\end{align}

\textbf{Current best estimate:} At time $t \geq 1$, the algorithm computes its current best estimate $\hat{\sigma}(t)$ as
\begin{align}
    \hat{\sigma}(t) = \argmin_{\sigma \in \mc{C}} g_{\hat{P}(t)}^{\sigma}(w(t)), \label{eq:sigmahat}
\end{align}
where $g_P^{\sigma}(w)$ is defined in \eqref{eq:gsV}. 
Ties in \eqref{eq:sigmahat} are broken arbitrarily.

\textbf{Computation of the optimal allocation:} The maximizer $w^*$ that achieves \eqref{eq:TstarP} acts as an oracle and indicates that if possible, a good algorithm pulls arms according to the allocation given by $w^*$. Track-and-Stop-type algorithms work according to the principle that at time $t$, the algorithm can efficiently compute the 
empirically optimal allocation with respect to the lower bound in \thmref{thm:lower}. Here, it is given by
\begin{align}
    w^*(t) &= \argmax_{w \in \Sigma_{K}} G_{\hat{P}(t-1)}^{\hat{\sigma}(t-1)}(w) \label{eq:wstar} \\
    &= \argmax_{w \in \Sigma_{K}} \min_{\sigma' \in \mc{C} \setminus \{\hat{\sigma}(t-1)\}} g_{\hat{P}(t-1)}^{\sigma'}(w).
\end{align}
Track-and-Stop-type algorithms ensure that the empirical allocation resulting from the algorithm, $w(t)$, approaches $w^*(t)$ as $t$ grows without bound. The latter is achieved via tracking methods such as C- and D-tracking proposed in \cite{garivier2016} and P-tracking proposed in \cite{tabata2023}. 

Efficiently and accurately computing \eqref{eq:wstar} is a challenging task in itself. See \secref{sec:discdifficult} for further discussion. In the following, we describe how we adapt the \textsf{FWS} algorithm from \cite{wang2021fast} to approximately compute \eqref{eq:wstar}. For $r \in (0, 1)$, we define the $r$-subdifferential subspace $H_{G_P^{\sigma}}(w, r)$ \cite{wang2021fast} as
\begin{align}
    H_{G_P^{\sigma}}(w, r) \triangleq \mr{co}( \nabla \, g_P^{\sigma'}(w) \colon \sigma' \neq \sigma, g_P^{\sigma'}(w) < G_P^{\sigma}(w) + r ),
\end{align}
where $\mr{co}(\cdot)$ denotes the convex hull of a set. In \cite[Lemma~10]{wang2021fast}, it is shown that $H_{G_P^{\sigma}}(w, r) \subseteq \partial_r G_P^{\sigma}(w)$, where the $r$-subdifferential of a concave function $f \colon \mc{A} \to \mathbb{R}$ with a compact domain $\mc{A} \subseteq \mathbb{R}^d$ is defined as 
\begin{align}
    \partial_r f(x) &\triangleq \{h \in \mathbb{R}^d \colon f(y) < f(x) + \langle y - x, h \rangle + r \notag \\
    &\quad \text{ for all } y \in \mc{A} \}. \label{eq:rsubdiff}
\end{align}
Notice that computing $H_{G_P^{\sigma}}(w, r)$ is simpler than computing $\partial_r G_P^{\sigma}(w)$ since $H_{G_P^{\sigma}}(w, r)$ requires computing the gradient only at finitely many points rather than a neighborhood of $w$.

Let $\{r_t\}_{t \geq 1}$ be a sequence that approaches zero sufficiently quickly. For concreteness, we set $r_t = t^{-4/5}$. The \textsf{FWS} update at time $t$ is given by
\begin{align}
    z(t) &= \argmax_{z \in \Sigma_{K}} \min_{h \in H_{G_{\hat{P}(t-1)}^{\hat{\sigma}(t-1)}}(x(t-1), r_t)} \langle z - x(t-1), h \rangle \label{eq:game}\\
    x(t) &= \left(1 - \frac{1}{t}\right) x(t-1) + \frac{1}{t} z(t) = \frac{1}{t} \sum_{s = 1}^t z(s). \label{eq:FWSx}
\end{align}
Since $ H_{G_P^{\sigma}}(w, r)$ is the convex hull of finitely many points, in \cite[Appendix~H]{wang2021fast}, the maximin problem is viewed as a solution to a zero-sum game, which then is converted into a linear program. Let $\mc{B} \subseteq \mc{C} \setminus \{\hat{\sigma}(t-1)\}$ be such that 
\begin{align}
    H_{G_{\hat{P}(t-1)}^{\hat{\sigma}(t-1)}}(x(t-1), r_t) = \mr{co}( \{ \nabla \, g_{\hat{P}(t-1)}^{\sigma'}(x(t-1)) \}_{\sigma' \in \mc{B}}).
\end{align}
The pay-off matrix of the game, $\mathsf{M} \in \mathbb{R}^{K \times |\mc{B}|}$, is defined as
\begin{align}
    \mathsf{M}_{k, \sigma'} \triangleq \langle e_k - x(t-1), \nabla g_P^{\sigma'}(x(t-1)) \rangle 
\end{align}
for $k \in [K], \sigma' \in \mc{B}$.
Then the linear program that solves \eqref{eq:game} is given by
\begin{align}
    &\max_{z \in \Sigma_{K}, u \in \mathbb{R}} u \notag \\
    &\text{s. t. } \quad \quad (z^\top \mathsf{M})_{\sigma'} \geq u \quad \text{for } \sigma' \in \mc{B}.  \label{eq:LPopt}
\end{align}

\textbf{Sampling strategy:} As in \cite{garivier2016, wang2021fast}, we employ a tracking strategy that has a forced exploration component.
Define the fixed forced exploration indices as
\begin{align}
    \mc{I}_{\mr{f}} \triangleq \{t \in \mathbb{N} \colon \lceil \sqrt{t} \log t \rceil = \lceil \sqrt{t+1} \log(t+1) \rceil - 1\}. \label{eq:Ifdef}
\end{align}
It holds that 
\begin{align}
    \lceil \sqrt{t} \log t \rceil \leq \lvert \mc{I}_{\mr{f}} \cap [t] \rvert \leq \lceil \sqrt{t} \log t \rceil + 1 \label{eq:If}
\end{align}
for $t \in \mathbb{N}$.

\subsubsection{C-tracking}
We employ the C-tracking method from \cite{garivier2016} to track the \textsf{FWS} estimate in each round.  

Let $\tilde{z}(t) \in \Sigma_{K}$ be a sequence whose accumulation is  tracked at time $t$. The C-tracking method pulls the arm $A_t \in [K]$ at time $t$ according to the rule
\begin{align}
    A_t = \argmax_{i \in [K]} \left( \left(\sum_{s = 1}^{t} \tilde{z}_i(s) \right)- N_{i}(t-1) \right), \label{eq:Atrule}
\end{align}
where $N(0) = \bs{0}$ and $N(t) = N(t-1) + e_{A_t}$. If there exist multiple arms that achieve the maximum, then the decision maker chooses one of them arbitrarily. 

\subsubsection{Uniform sampling phase} To ensure  that each arm is pulled once up to time $K$, we set
\begin{align}
    \tilde{z}(t) = \frac{1}{K} \bs{1}, \quad t \in [K]. \label{eq:zcases}
\end{align}

\subsubsection{\textsf{FWS} phase with forced exploration} 
Initialize $\tilde{x}(K) = \frac{1}{K} \bs{1}$. Starting from the first time after the uniform sampling phase ends, i.e., $t = K + 1$, we modify the \textsf{FWS} update in \eqref{eq:game}--\eqref{eq:FWSx} as
\begin{align}
    \tilde{z}(t) = 
    \argmax_{z \in \Sigma_{K}} \min_{h \in H_{G_{\hat{P}(t-1)}^{\hat{\sigma}(t-1)}}(\tilde{x}(t-1), r_t)} \langle z - \tilde{x}(t-1), h \rangle \label{eq:ztilde}
\end{align}
if $t \notin \mc{I}_{\mr{f}}$, and
\begin{align}
    \tilde{z}(t) = \frac{1}{K} \bs{1} 
\end{align}
if $t \in \mc{I}_{\mr{f}}$, where
\begin{align}
    \tilde{x}(t) = \frac{1}{t}\sum_{s = 1}^t \tilde{z}(s). \label{eq:xtildedef}
\end{align}


\textbf{Stopping rule and recommendation:} 
The stopping criterion is based on the statistics
\begin{align}
    Z(t) &= t \, G_{\hat{P}(t)}^{\hat{\sigma}(t)}(w(t)) \\
    &= t \, \min_{\sigma \neq \hat{\sigma}(t)} g_{\hat{P}(t)}^{\sigma}(w(t)). \label{eq:Zt}
\end{align}
The right-hand side of \eqref{eq:Zt} is equal to $t$ times the second smallest $g_{\hat{P}(t)}^{\sigma}(w(t))$ at time $t$.
Let
\begin{align}
    \tilde{K} \triangleq \max_{\sigma \in \mc{C}} \left \lvert \bigcup_{m \in [M]} \mc{A}_m^{\sigma} \right \rvert
\end{align}
be the maximum number of arms that belong to a cluster. The algorithm stops at time
\begin{align}
    \tau \triangleq \inf\{t \geq 1 \colon Z(t) \geq \beta(t, \delta)\},  \label{eq:taudef}
\end{align}
where 
\begin{align}
    \beta(t, \delta) &= \log \frac{1}{\delta} + (M|\mc{X}| + \tilde{K} + 2) \log(t + 1) \notag \\
    &\quad + \log \left( \frac{\pi^2}{6} - 1 \right) \label{eq:betadef}
\end{align}
is the threshold of our algorithm. We prove in \secref{sec:correctness}, below, that the threshold in \eqref{eq:betadef} guarantees the $\delta$-correctness of the algorithm.
The algorithm recommends the hypothesis $\hat{\sigma}(\tau)$. The pseudo-code of our algorithm appears in Algorithm~\ref{algo}, below. See \secref{sec:discAlgo} for the comparison of our algorithm with the other PE algorithms in the literature. 

\begin{algorithm}[!htbp] 
 \caption{\textsf{TaS-FW}}
 \label{algo}
 \begin{algorithmic}[1]
 \renewcommand{\algorithmicrequire}{\textbf{Input:}}
 \renewcommand{\algorithmicensure}{\textbf{Output:}}
 \REQUIRE Target error probability $\delta \in (0, 1)$, the collection of hypotheses $\mc{C}$
 \\ \textit{Initialization}: Sample each arm $i \in [K]$ once, initialize $\tilde{x}(K) = \frac{1}{K} \boldsymbol{1}$ and $N(K) = (1, \dots, 1)$, and update $\hat{P}(K)$, $\hat{\sigma}(K)$, and $Z(K)$. For $t \in \mathbb{N}$, set $r_t = t^{-4/5}$. \\ $t \gets K$
  \WHILE{$Z(t) < \beta(t, \delta)$}
  \STATE $t \gets t + 1$
  \IF {$t \in \mc{I}_{\mr{f}}$}
  \STATE $\tilde{z}(t) \gets \frac{1}{K} \bs{1}$
  \ELSIF {$t \notin \mc{I}_{\mr{f}}$}
  \STATE $\tilde{z}(t) \gets 
    \argmax\limits_{z \in \Sigma_{K}} \min\limits_{h \in H_{G_{\hat{P}(t-1)}^{\hat{\sigma}(t-1)}}(\tilde{x}(t-1), r_t)} \langle z - \tilde{x}(t-1), h \rangle$
  \ENDIF
  \STATE $\tilde{x}(t) \gets \left(1 - \frac{1}{t}\right) \tilde{x}(t-1) + \frac{1}{t} \tilde{z}(t)$
  \STATE Sample the arm $A_t \gets \argmax \limits_{i \in [K]} \left( t \tilde{x}_i(t) - N_{i}(t-1) \right)$
  \STATE Update $N(t) \gets N(t-1) + e_{A_t}$ and the empirical problem instance $\hat{P}(t)$ in \eqref{eq:hatPdef}
  \STATE  $\hat{\sigma}(t) \gets \argmin \limits_{\sigma \in \mc{C}} g_{\hat{P}(t)}^{\sigma}(w(t))$
  \ENDWHILE
  \ENSURE  $\hat{\sigma}(t)$
 \end{algorithmic} 
 \end{algorithm}



\thmref{thm:upper}, below, which is the main result of our paper, is an instance-dependent achievability bound that gives an upper bound on the average stopping time of our algorithm \textsf{TaS-FW} described above.
\begin{theorem}\label{thm:upper}
    Let $\mc{C}$ be a problem that Assumption~\ref{as:finer} holds. For any problem instance $P \in \Lambda$, as $\delta \to 0^+$, our algorithm \textsf{TaS-FW} described in Section \ref{sec:TAS} is $\delta$-correct and achieves
    \begin{align}
        \E{\tau} \leq   \frac{\log\frac{1}{\delta}}{T^*(P)} \left( 1 + O\left( \left(\log \frac{1}{\delta} \right)^{-1/4} \sqrt{\log \log \frac{1}{\delta}} \right) \right). \label{eq:Etauupper}
    \end{align}
\end{theorem}
\begin{IEEEproof}
   See \secref{sec:proof}.
\end{IEEEproof}

Comparing the lower bound in \thmref{thm:lower} and the upper bound in \thmref{thm:upper}, we see that in both bounds, the leading term on the right hand side is $\frac{\log \frac{1}{\delta}}{T^*(P)}$. This shows that our algorithm \textsf{TaS-FW} achieves an average stopping time that is asymptotically optimal as the error probability $\delta$ goes to 0. \thmref{thm:upper} also shows that the additive second-order term in the upper bound scales as $O\left( \left(\log \frac{1}{\delta} \right)^{3/4} \sqrt{\log \log \frac{1}{\delta}} \right)$. We carefully select the design parameters such as the forced exploration frequency in \eqref{eq:If} to optimize the scaling of the second-order term in the upper bound. The coefficient of the $O(\cdot)$ term in \eqref{eq:Etauupper} can be found in \eqref{eq:T0asymsec}. The expression on the right-hand side of \eqref{eq:Etauupper} is derived by analyzing the non-asymptotic bound in \eqref{eq:ETbound} as $\delta \to 0^+$.

The lower bound in \thmref{thm:lower} states that $\E{\tau} \geq \frac{\log \frac{1}{\delta}}{T^*(P)} + \Theta(1)$. It remains an open problem to close the gap between the lower and upper bounds in the second-order term.

\section{Experimental Results}\label{sec:experiments}
In this section, we numerically evaluate the average number of arm pulls of our algorithm for each of the example problems in the introduction. In all three examples, our empirical findings support the theoretical results in Theorems~\ref{thm:lower}--\ref{thm:upper}.
The number of independent trials is 300 for Examples 1 and 2 and 100 for Example 3. For each of the experiments, the empirical error probability is equal to zero.

\subsection{Example 1: Matching pairs with two groups of arms}
In this example, the number of arms is $K = 6$, the number matching pairs is $M = 2$, the indices of nominal and candidate arms are $\{1, 2\}$ and $\{3, 4, 5, 6\}$, respectively. We consider two problem instances with the alphabet sizes $|\mc{X}| = 3$ and $|\mc{X}| = 5$. 
The underlying problem instance is $P = (P_1, \dots, P_6)$. For $|\mc{X}| = 3$, $P_1 = P_3 = (0.1, 0.1, 0.8)$, $P_2 = P_4 = (0.4, 0.4, 0.2)$, $P_5 = (0.5, 0.05, 0.45)$, and $P_6 = (0.1, 0.8, 0.1)$. For $|\mc{X}| = 5$, $P_1 = P_3 = (0.1, 0.1, 0.6, 0.1, 0.1)$, $P_2 = P_4 = (0.2, 0.2, 0.2, 0.2, 0.2)$, $P_5 = (0.4, 0.05, 0.1, 0.05, 0.4)$, and $P_6 = (0.1, 0.6, 0.1, 0.1, 0.1)$.
Hence, the true hypothesis associated with $P$ is
\begin{align}
    \sigma_P = \{\{1, 3\}, \{2, 4\}\}.
\end{align}

\begin{figure}[!htbp]
\center
\includegraphics[width=1\linewidth]{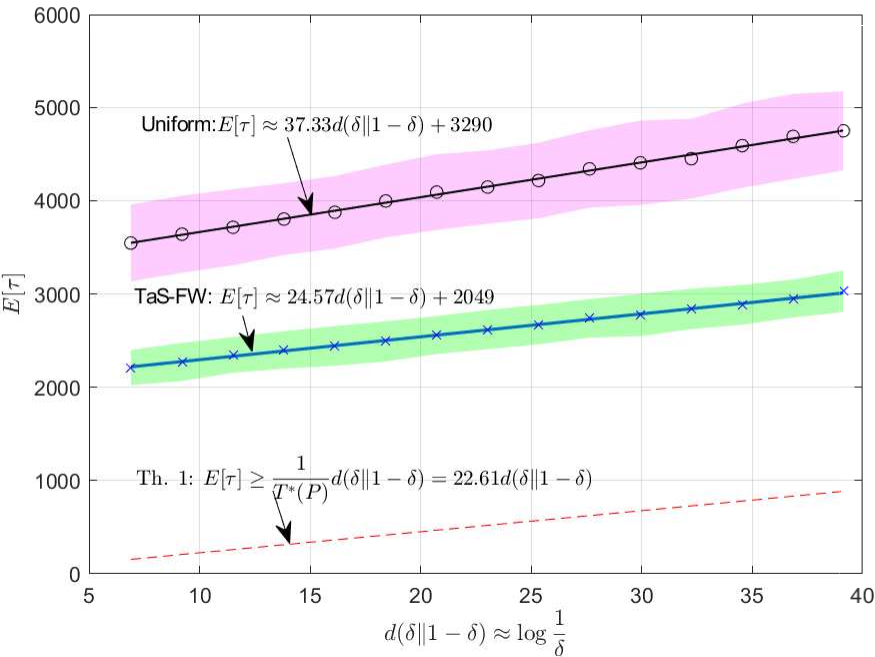}
\caption{Example 1: Matching pairs with two groups of arms with $|\mc{X}| = 3$.}
\label{fig:example1}
\end{figure}

\begin{figure}[!htbp]
\center
\includegraphics[width=1\linewidth]{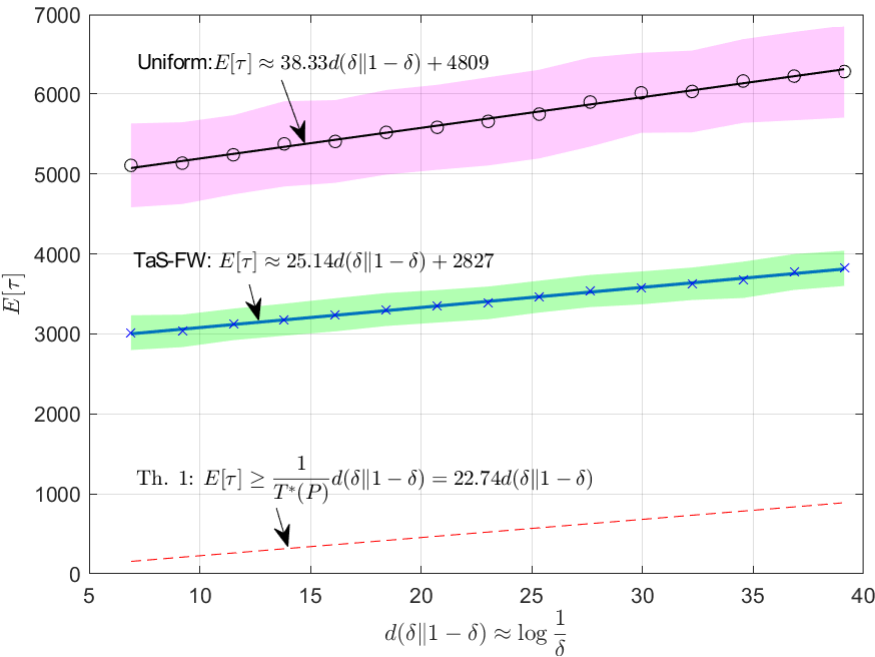}
\caption{Example 1: Matching pairs with two groups of arms with $|\mc{X}| = 5$.}
\label{fig:example1.X5}
\end{figure}

In \figref{fig:example1}--\ref{fig:example1.X5}, we demonstrate the empirical mean and standard deviation of our algorithm \textsf{TaS-FW} and compare its performance with that of the \textsf{Uniform} sampling strategy that samples arms uniformly at random at each time step and employs the same stopping rule (in \eqref{eq:taudef}) as \textsf{TaS-FW}. The shaded regions indicate the values within 1-standard-deviation from the empirical mean.  
We also compare the average number of pulls from these experiments with the lower bound in \thmref{thm:lower}. For all $\delta \in [10^{-17}, 10^{-3}]$, \textsf{TaS-FW} outperforms \textsf{Uniform}. For $|\mc{X}| = 3$ (and for $|\mc{X}| = 5$), the slope of the linear regression applied to the empirical average number of pulls of \textsf{TaS-FW} with respect to $d(\delta \| 1 - \delta) \approx \log \frac{1}{\delta}$ is $24.57$ (25.14), which is reasonably close to the hardness parameter $\frac{1}{T^*(P)} = 22.61$ (22.74) that appears in the lower bound in \thmref{thm:lower}. On the other hand, the slope associated with \textsf{Uniform} is 37.33 (38.33), which is much higher than the theoretical lower bound.

\subsection{Example 2: Odd arm identification}
In this example, we consider the standard odd arm identification problem with only a single odd arm. The goal is to identify the index of the odd arm among $K$ arms. 
The number of arms is $K = 7$, the number of clusters is $M = 1$, and the size of the cluster is $K - 1 = 6$. The underlying problem instance is $P = (P_1, \dots, P_7)$, where $P_i = (0.1, 0.1, 0.8)$ for $i \in [6]$, and $P_7 = (0.6, 0.2, 0.2)$. Hence, arm 7 is the odd arm, equivalently,
\begin{align}
    \sigma_P = \{\{1, 2, 3, 4, 5, 6\}\}. 
\end{align}
\begin{figure}[!htbp]
\center
\includegraphics[width=1\linewidth]{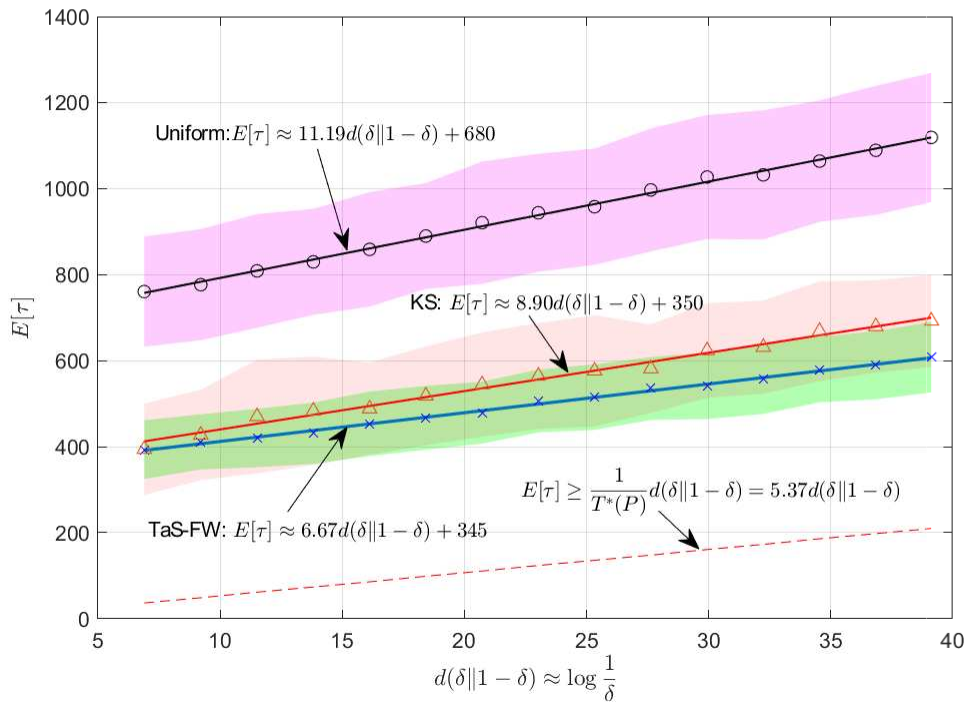}
\caption{Example 2: Odd arm identification.}
\label{fig:example2}
\end{figure}

In \figref{fig:example2}, we demonstrate the empirical mean and standard deviation of the number of arm pulls for \textsf{TaS-FW}, \textsf{Uniform}, and Karthik and Sundaresan's algorithm (labeled as \textsf{KS}) in \cite{karthik2019}, and we compare those with the lower bound. The uniform sampling parameter of $\textsf{KS}$ is set to 0.1, i.e., with probability 0.1, the next arm is selected uniformly at random. The \textsf{KS} algorithm was originally designed for Markov arms; our problem is a special case of \cite{karthik2019} with a single state.  
Both \textsf{TaS-FW} and \textsf{KS} (for any uniform sampling parameter in $(0, 1)$) are asymptotically optimal algorithms. For this particular problem instance, we observe that for all $\delta \in [10^{-17}, 10^{-3}]$, \textsf{TaS-FW} empirically outperforms \textsf{KS} and \textsf{Uniform} in terms of the average number of arm pulls. The slope of the linear regression line for \textsf{TaS-FW} is 6.67, which is the smallest among the three algorithms shown. The hardness parameter $\frac{1}{T^*(P)}$ is reported as 5.37.

\subsection{Example 3: $N$-ary clustering of $K$ arms}
In this example, we consider $N$-ary clustering of $K$ arms introduced in \cite{yang2024clustering}. There are $K$ arms that are partitioned into $N$ groups where each group can have any size greater than or equal to 1. The number of arms is $K = 6$, and the number of clusters is $N = 3$. The problem instance is $P = (P_1, \dots, P_7)$, where $P_1 = P_2 = (0.6, 0.2, 0.2)$, $P_3 = P_4 = (0.25, 0.7, 0.05)$, and $P_5 = P_6 = (0.05, 0.05, 0.90)$. Hence, the true hypothesis associated with $P$ is 
\begin{align}
    \sigma_P = \{\{1, 2\}, \{3, 4\}, \{5, 6\}\}. \label{eq:sigmaPex3}
\end{align}
\begin{figure}[!htbp]
\center
\includegraphics[width=1\linewidth]{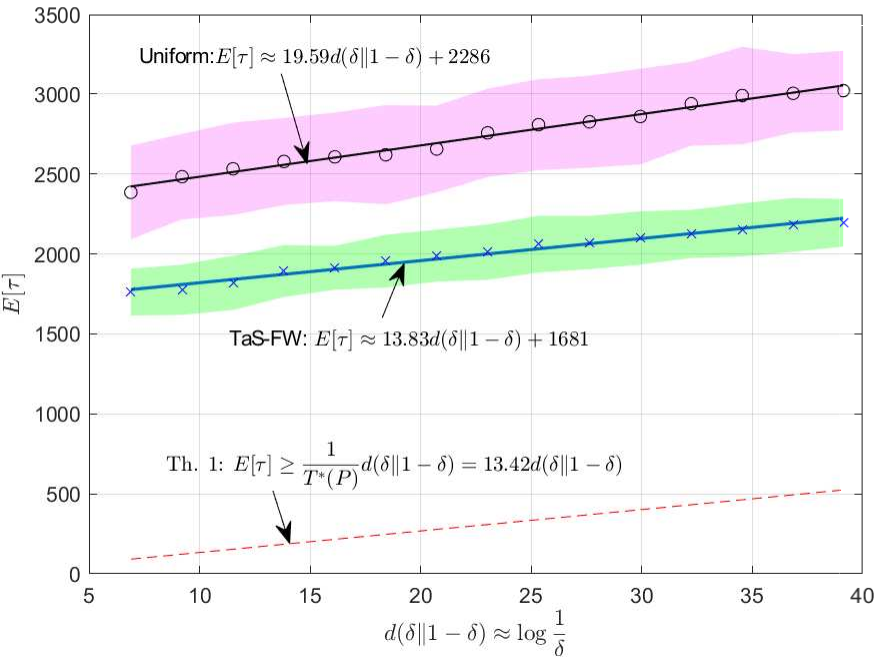}
\caption{Example 3: $N$-ary clustering of $K$ arms.}
\label{fig:example3}
\end{figure}

In \figref{fig:example3}, we demonstrate the empirical mean and standard deviation of the number of arm pulls for \textsf{TaS-FW} and \textsf{Uniform}, and we compare those with the lower bound. For all $\delta \in [10^{-17}, 10^{-3}]$, \textsf{TaS-FW} outperforms \textsf{Uniform}. The slope associated with \textsf{TaS-FW} (13.83) is only 3\% larger than the hardness parameter $\frac{1}{T^*(P)} = 13.42$, while the slope associated with \textsf{Uniform} is much higher (19.59).

\subsection{Example 3: $N$-ary clustering of $K$ arms with Gaussian Distribution} \label{sec:Gaussian}

Consider the problem in Example 3 where arm $i$ follows $\mathcal{N}(\mu_i, 1)$ with an unknown mean $\mu_i$ and a known variance~$1$.
Although our achievability bound in \thmref{thm:upper} does not provide a theoretical guarantee for the clustering problems with Gaussian distributed arms, \textsf{TaS-FW} is applicable to such a scenario with the following modifications:
\begin{itemize}
    \item The empirical distribution $\hat{P}_i(t) $ is redefined as
    \begin{align}
        \hat{P}_i(t) = \mc{N}(\bar{X}_i(t), 1),
    \end{align}
    where $\bar{X}_i(t) = \frac{1}{N_i(t)} \sum_{n = 1}^t X_{n, A_n} 1\{A_n = i\}$ is the empirical reward of arm $i$ at time $t$.
    \item We modify the threshold $\beta(t, \delta)$ in \eqref{eq:betadef} by setting $|\mc{X}| = 2$. Alternatively, we can use the threshold given in \cite[eq.~(8)]{yang2024clustering}. This is justified because $Z(t)$ in Line 12 of \textsf{BOC} in \cite{yang2024clustering} is an approximation of the statistics $Z(t)$ in \eqref{eq:Zt}.
\end{itemize}

We conduct an experiment to compare the performance of \textsf{TaS-FW} with that of Yang et al.'s \textsf{BOC} algorithm \cite{yang2024clustering}. For \textsf{TaS-FW}, we use two different threshold functions mentioned above. For the problem instance, the number of arms is set to $K = 6$, and the number of groups is $N = 3$. The underlying problem instance is $P = (P_1, \dots, P_6)$, where $P_1 = P_2 = \mc{N}(0, 1)$, $P_3 = P_4 = \mc{N}(5, 1)$, and $P_5 = P_6 = \mc{N}(-2, 1)$. Hence, $\sigma_P$ is as in \eqref{eq:sigmaPex3}.

\begin{figure}[!htbp]
\center
\includegraphics[width=1\linewidth]{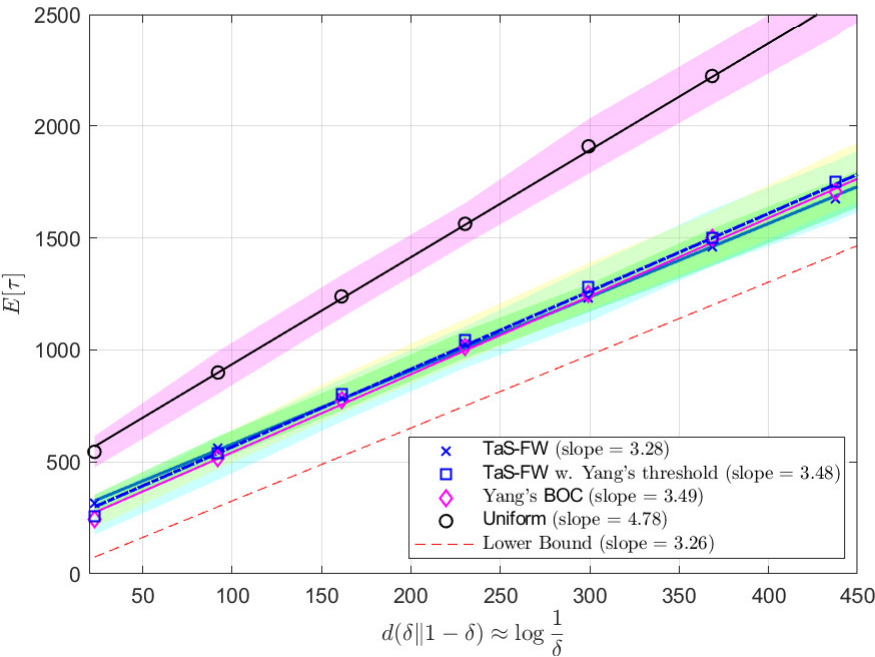}
\caption{Example 3: $N$-ary clustering of $K$ arms with Gaussian distribution.}
\label{fig:example3Gauss}
\end{figure}

In Fig.~\ref{fig:example3Gauss}, we demonstrate the empirical mean and standard deviation of the number of arm pulls for \textsf{TaS-FW}, \textsf{BOC}, and \textsf{Uniform}. To get a more accurate estimate of the slopes, we set the target error probability to smaller values: $\delta \in [10^{-200}, 10^{-10}]$. The performances of \textsf{TaS-FW} and \textsf{BOC} are barely distinguishable. The slope associated with \textsf{TaS-FW} with our threshold (3.28) is the smallest among the compared algorithm and only 1\% larger than the slope of the lower bound (3.25).  The slopes associated with \textsf{TaS-FW} with Yang et al.'s threshold and \textsf{BOC} are 3.48 and 3.49, respectively, which are roughly 7\% away from the lower bound. As expected, \textsf{Uniform} performs much worse (4.78) than both \textsf{TaS-FW} and \textsf{BOC}. 

Despite having slightly better empirical performance on the given example, as shown in Table~\ref{table:comp} below, the computation time of \textsf{TaS-FW} is roughly 5 times larger than that of \textsf{BOC}. This is primarily due to the fact that  \textsf{BOC} exploits the sum-of-squares structure of the objective function to leverage a version of the $K$-means algorithm to solve the inner infimum in~\eqref{eq:3opt}  efficiently. However, for discrete alphabets, such a simplification does not appear to be possible as the objective function is not of the sum-of-squares form.

In order to get a theoretical guarantee for the Gaussian case, one needs to extend the results in \secref{sec:convFWS} and the large deviation bound in \lemref{lem:concent} to Gaussian distributions. This is left for future work.

\subsection{Computation Times} \label{sec:comp}
In Table~\ref{table:comp}, we demonstrate the computation times of the compared algorithms for Examples 1--3 above. We did each experiment on Google Colab. 

The primary computational bottleneck of \textsf{TaS-FW} arises from solving Line 6 of Algorithm~\ref{algo} to determine the allocation of arm pulls, which reduces to the linear program in \eqref{eq:LPopt}.
The linear program in \eqref{eq:LPopt} has $K + 1$ variables and $B$ constraints, where $K$ is the number of arms, and $B$ is the number of hypotheses in the $r$-subdifferential subspace.
In the worst case, $B$ can approach the size of the entire hypothesis set $\mathcal{C}$, making the time complexity of the algorithm as high as $O(|\mathcal{C}|^{2.5})$.
Even in more favorable cases where $B$ is small relative to $|\mc{C}|$, the complexity is at least $\Omega(|\mc{C}|)$ since obtaining the current best estimate $\hat{\sigma}(t)$ in Line 11 of Algorithm~\ref{algo} requires to compute the score function for all $|\mc{C}|$ hypotheses. 

In Example 3, for \textsf{TaS-FW}, we observe that the computation time quickly increases as $K$ grows. This is because the number of hypotheses $|\mc{C}|$ grows exponentially with $K$ for Example 3. On the other hand, Yang et al.'s \textsf{BOC} algorithm \cite{yang2024clustering} is much faster because they do not enumerate all hypotheses in $\mc{C}$ but instead they exploit the structure of this specific example together with the KL divergence between two Gaussian distributions being related to the Euclidean distance between the mean vectors of the distributions. 

We found that changes in the alphabet size and distribution type (finite alphabet or Gaussian) do not have a significant effect on the computation time. This is because the alphabet size mainly affects the KL divergence computation, which has low complexity relative to the linear program.


\begin{table}[htbp!] 
\centering
\caption{Computation times of several algorithms for Examples 1--3}
\label{table:comp}
\begin{tabular}{c c c c c c c}
\hline
Example \# & $K$ & $M$ & $|\mc{C}|$ & $|\mc{X}|$ & Algo. & Comp. time \\ & & & & & & (sec/1000 steps) \\
\hline
1 & 6 & 1 & 5 & 2 & \textsf{TaS-FW} & 7.2 \\
1 & 6 & 1 & 5 & 5 & \textsf{TaS-FW} & 7.4 \\
1 & 6 & 1& 5 & 10 & \textsf{TaS-FW} & 7.6\\
1 & 6 & 2& 12 & 5 & \textsf{TaS-FW} & 16.1 \\
1 & 12 & 2& 90 & 5 & \textsf{TaS-FW} & 109.8\\
1 & 12 & 5& 2520 & 5 & \textsf{TaS-FW} & 6634.1\\
\hline
2 & 7 & - & 6& 3 & \textsf{TaS-FW} & 14.3 \\
2 & 7 & - & 6& 3 & \textsf{KS} \cite{karthik2019} & 54.3 \\
2 & 14 & - & 13& 3 & \textsf{TaS-FW} & 43.4 \\
2 & 14 & - & 13& 3 & \textsf{KS} \cite{karthik2019} & 81.6 \\
2 & 28 & - & 27& 3 & \textsf{TaS-FW} & 146.8 \\
2 & 28 & - & 27& 3 & \textsf{KS} \cite{karthik2019} & 204.2 \\
\hline
3 & 6 & 3& 90 & Gauss. & \textsf{TaS-FW} & 152.9 \\
3 & 6 & 3& 90 & Gauss. & \textsf{BOC} \cite{yang2024clustering} & 30.5 \\
3 & 7 & 3& 301 & Gauss. & \textsf{TaS-FW} & 505.9 \\
3 & 7 & 3& 301 & Gauss. & \textsf{BOC} \cite{yang2024clustering} & 33.4 \\
3 & 8 & 3& 966 & Gauss. & \textsf{TaS-FW} & 1413.6 \\
3 & 8 & 3& 966 & Gauss. & \textsf{BOC} \cite{yang2024clustering} & 34.4 \\
3 & 6 & 3& 90 & 3 & \textsf{TaS-FW} & 210.5 \\
3 & 7 & 3& 301 & 3 & \textsf{TaS-FW} & 777.8\\
3 & 8 & 3& 966 & 3 & \textsf{TaS-FW} & 2598.9\\
\hline
\end{tabular}
\end{table}

\section{Proof of \thmref{thm:upper}} \label{sec:proof}
The roadmap to prove \thmref{thm:upper} is as follows. In \secref{sec:convCtrack}, we present a lemma and its corollary that analyzes the C-tracking method in \eqref{eq:Atrule}. In \secref{sec:convFWS}, we first present several auxiliary lemmas on the properties (e.g., Lipschitzness and the curvature function) of the functions $g_P^{\sigma}(w)$ and $G_P^{\sigma}(w)$ in \eqref{eq:gsV}--\eqref{eq:GsV}. We then use these lemmas to derive a non-asymptotic bound, under a suitably chosen high probability event, on the gap between the quantity $T^*(P) = G_P^{\sigma_P}(w^*)$ in \eqref{eq:TstarP}, which appears in the lower bound in \thmref{thm:lower}, and its empirical counterpart $G_P^{\sigma_P}(w(t))$ that is evaluated at the allocation $w(t)$ resulting from our algorithm. In \secref{sec:deltacorrect}, we present a concentration inequality on the function $g_{\hat{P}(t)}^{\sigma_P}(w)$ evaluated at the empirical problem instance $\hat{P}(t)$ and the true hypothesis $\sigma_P$. We then apply this result to bound the error probability of \textsf{TaS-FW}. Finally, we use the results in Sections~\ref{sec:convCtrack}--\ref{sec:convFWS} to derive an upper bound on the average stopping time of \textsf{TaS-FW}.

\subsection{Convergence of the C-tracking Method} \label{sec:convCtrack}
The following result from \cite[Lemma~15]{garivier2016} controls the gap between the number of pulls for arm $i$ up to time $t$, $N_i(t)$, and the tracked sequence $\sum_{s = 1}^t \tilde{z}_i(s)$ in \eqref{eq:Atrule}.
\begin{lemma}[\kern-1ex{\cite[Lemma~15]{garivier2016}}]\label{lem:ctrack}
    Fix an integer $K \geq 1$. Let $\{\tilde{z}(s)\}_{s\geq 1}$ be a sequence on $\Sigma_K$. Define $N(0) = \boldsymbol{0}$, and for every $t \geq 1$,
    \begin{align}
        A_t \in \argmax_{i \in [K]} \left(\left( \sum_{s = 1}^t \tilde{z}_i(s) \right) - N_i(t-1)\right),
    \end{align}
    and $N(t) = N(t-1) + e_{A_t} \in \mathbb{N}^K$. Then, for all $t \geq 1$ and $i \in [K]$, 
    \begin{align}
        \left \lvert N_i(t) - \sum_{s = 1}^t \tilde{z}_i(s) \right \rvert \leq K - 1.
    \end{align}
\end{lemma}
The next result lower bounds the number of pulls for each arm up to time $t$.
\begin{corollary} \label{cor:wtlower}
    Our \textsf{TaS-FW} algorithm guarantees that for all $i \in [K]$ and $t \in \mathbb{N}$, it holds that
    \begin{align}
        N_i(t) \geq \frac{\sqrt{t} \log t}{K} - K + 1.
    \end{align}
\end{corollary}
\begin{IEEEproof}
   By the procedure in \eqref{eq:ztilde}--\eqref{eq:xtildedef} and the bound in \eqref{eq:If}, we have 
    \begin{align}
        N_i(t) &\geq \sum_{s = 1}^t \tilde{z}_i(s) - (K - 1) \label{eq:Nz} \\
        &\geq \frac{1}{K} |\mc{I}_{\mr{f}} \cap [t] | \\
        &\geq \frac{1}{K} \lceil \sqrt{t} \log t \rceil - (K - 1)\label{eq:lastK} \\
        &\geq \frac{\sqrt{t} \log t}{K} - K + 1, 
    \end{align}
    where \eqref{eq:Nz} applies \lemref{lem:ctrack}, and \eqref{eq:lastK} applies \eqref{eq:If}.
\end{IEEEproof}

\subsection{Convergence of the \textsf{FWS} Algorithm} \label{sec:convFWS}
Recall from \eqref{eq:TPsigma} and \eqref{eq:Tmax} that $T^*(P) = G_P^{\sigma_P}(w^*)$ where $w^* \in \Sigma_K$ is the maximizer of the function $G_P^{\sigma_P}(w)$. For $P \in \Lambda$, for brevity, we denote $G_P \triangleq G_P^{\sigma_P}$.

We define the optimality gaps with respect to the tracked sequence $\tilde{x}(t)$ in \eqref{eq:xtildedef} and the empirical allocation vector $w(t)$ in \eqref{eq:wt} as
\begin{align}
    \Delta_t &\triangleq G_P(w^*) - G_P(\tilde{x}(t)). \label{eq:Deltatdef} \\
    \tilde{\Delta}_{t} &\triangleq G_P(w^*) - G_P(w(t)). \label{eq:Deltattilde}
\end{align}
In \cite{wang2021fast}, a crucial step to bound the average stopping time of the \textsf{FWS} algorithm is to upper bound $\tilde{\Delta}_t$ under a high probability event on the empirical problem instances $\hat{P}(t)$. To do this, for a general PE problem, they assume that the functions $g_P^{\sigma}(w)$ and $G_P^{\sigma}(w)$ are $L$-Lipschitz in $w$ and $E$-Lipschitz in $P$, where $L$ and $E$ are some positive finite constants. They also assume that there exists a finite constant $D$ such that the curvature constant of $g_P^{\sigma}(\cdot)$ restricted on the set $\Sigma_K^{\gamma}$ (see \eqref{eq:Cfdef} for its definition) is bounded by $\frac{D}{\gamma}$ for all $\gamma \in (0, 1)$. Then, they proceed to bound $\tilde{\Delta}_t$ under these assumptions. In Lemmas~\ref{lem:gradg}--\ref{lem:curvature} below, we verify \cite[Assumptions~2~and~3]{wang2021fast} for general clustering problems with arm distributions on a finite alphabet, and we explicitly derive the constants $L, D$, and $E$. In Lemmas~\ref{lem:GPoracle}--\ref{lem:maxminz}, we prove two auxiliary results on $\Delta_t$ using the properties of $g_P^{\sigma}(w)$ and $G_P^{\sigma}(w)$ derived in Lemmas~\ref{lem:gradg}--\ref{lem:curvature}. Finally, in Lemma~\ref{lem:T2gap}, we bound $\Delta_t$ under a high probability event on $\hat{P}(t)$, and in Corollary~\ref{cor:GP}, we bound $\tilde{\Delta}_t$ under the same high probability event. 

\lemref{lem:gradg}, below, shows the Lipschitzness of $g_P^{\sigma}(w)$ and $G_P^{\sigma}(w)$ in $w$.
\begin{lemma} \label{lem:gradg}
    Fix some hypothesis $\sigma \in \mc{C}$. For all $(w, P) \in (\mr{int}(\Sigma_{K}), \Lambda_{\sigma})$, and $\sigma' \neq \sigma$, $g_P^{\sigma'}(w)$ is continuously differentiable with respect to $w$, and
    \begin{align}
        \nabla_{w} \, g_P^{\sigma'}(w) = \sum_{m = 1}^M \sum_{i_m \in \mc{A}_m^{\sigma'}} D(P_{i_m} \| W_{m}) e_{i_m}\label{eq:gradw}
    \end{align}
    where
    \begin{align}
        W_{m} = \frac{\sum_{i_m \in \mc{A}_{m}^{\sigma'}} w_{i_m} P_{i_m}}{\sum_{i_m \in \mc{A}_{m}^{\sigma'}} w_{i_m}}.
    \end{align}
    Consequently, $g_P^{\sigma'}(w)$ and therefore $G_P^{\sigma}(w)$ are $L$-Lipschitz in $w$ with respect to the $\ell_{\infty}$-norm where
    \begin{align}
         L \triangleq \max\limits_{\substack{\sigma' \neq \sigma \\m \in [M]}} \max_{i_m, j_m \in \mc{A}_{m}^{\sigma'}} D(P_{i_m} \| P_{j_m}). \label{eq:L}
    \end{align}
\end{lemma}
\begin{IEEEproof}
See \appref{app:gradg}.
\end{IEEEproof}

\lemref{lem:gradLip}, below, shows the Lipschitzness of $\nabla_w g_P^{\sigma'}(w)$ in $w$ on the set $\Sigma_K^{\gamma}$ for some $\gamma \in (0, \frac{1}{K})$. 

\begin{lemma} \label{lem:gradLip}
Let $P$ be an instance in $\Lambda$. Define 
\begin{align}
    p_{\min} \triangleq \min_{i \in [K]} \min_{a \in \mc{X}} P_i(a), \label{eq:pmin}
\end{align}
which is positive by assumption.
Fix some $\gamma \in (0, \frac{1}{K})$ and $\sigma' \neq \sigma_P$. Recall that $\Sigma_K^{\gamma} \triangleq \{w \in \Sigma_K \colon w_i \geq \gamma \, \forall \, i \in [K]\}$. On the set $\Sigma_K^{\gamma}$, the gradient $\nabla_w g_P^{\sigma'}(w)$ is $\frac{D}{\gamma}$-Lipschitz in $w$ with respect to the $\ell_{\infty}$-norm, where
\begin{align}
    D \triangleq  \left( \max_{\sigma' \in \mc{C}} \max_{m \in [M]} |\mc{A}_m^{\sigma'}| \right) \frac{ |\mc{X}| (1 - p_{\min})}{4 p_{\min}}. \label{eq:D}
\end{align}
\end{lemma}
\begin{IEEEproof}
See \appref{app:gradLip}.
\end{IEEEproof}

The following result establishes the Lipschitzness of the functions $g_P^{\sigma}(w)$ and $G_P^{\sigma}(w)$ in $P$ with respect to the $\ell_{\infty}$-norm.
\begin{lemma} \label{lem:LipinP}
Let $P\in \mc{P}^{K}(\mc{X})$ with $\frac{1}{2} \geq p_{\min} = \min_{i \in [K]} \min_{a \in \mc{X}} P_k(a) > 0$. Let $\epsilon \in \left(0, \frac{p_{\min}}{2}\right)$. Then, for all $Q \in \mc{P}^K(\mc{X})$ such that $\norm{P - Q}_{\infty} \leq \epsilon$ and for all $w \in \mr{int}(\Sigma_K)$ and $\sigma \in \mc{C}$,
\begin{align}
    |g_P^{\sigma}(w) - g_Q^{\sigma}(w)| &\leq E \epsilon \\
    |G_P^{\sigma}(w) - G_Q^{\sigma}(w)| &\leq E \epsilon, \label{eq:GPLipP}
\end{align}
where
\begin{align}
    E \triangleq |\mc{X}| \log \frac{2 - p_{\min}}{p_{\min}}. \label{eq:E}
\end{align}
\end{lemma}
\begin{IEEEproof}
See \appref{app:LipinP}.
\end{IEEEproof}

The following result shows that the optimal value of the maximin problem in the \text{FWS} algorithm in \eqref{eq:game} is Lipschitz in $P$.  
\begin{lemma}[\kern-1ex{\cite[Th.~9]{wang2021fast}}] \label{lem:LipinFW}
    Let $P, Q \in \mc{P}^{K}(\mc{X})$ such that $\frac{1}{2} \geq p_{\min} = \min_{i \in [K]} \min_{a \in \mc{X}} P_i(a) > 0$. Let $\epsilon \in (0, \frac{p_{\min}}{2})$, and let $E$ be as in \eqref{eq:E}. Suppose that $\norm{P - Q}_{\infty} \leq \epsilon$. Fix any hypothesis $\sigma \in \mc{C}$. Then,
    \begin{align}
        &\left \lvert \max \limits_{z \in \Sigma_{K}} \min \limits_{h \in H_{G_P^{\sigma}(w, r)}} \langle z - w, h \rangle - \max \limits_{z \in \Sigma_{K}} \min \limits_{h \in H_{G_Q^{\sigma}(w, r)}} \langle z - w, h \rangle \right \rvert \notag \\
        &\leq E \epsilon  
    \end{align}
    for all $(w, r) \in \mr{int}(\Sigma_K) \times (0, 1)$, and
    \begin{align}
        \left \lvert \min \limits_{h \in H_{G_P^{\sigma}(w, r)}} \langle z - w, h \rangle - \min \limits_{h \in H_{G_Q^{\sigma}(w, r)}} \langle z - w, h \rangle \right \rvert \leq E\epsilon  
    \end{align}
    for all $(z, w, r) \in \Sigma_K \times \mr{int}(\Sigma_K) \times (0, 1)$.
    
    Consequently, for 
    \begin{align}
        z^* \in \argmax_{z \in \Sigma_K} \min \limits_{h \in H_{G_Q^{\sigma_P}}(w, r)} \langle z - w, h \rangle,
    \end{align}
    the above inequalities imply
    \begin{align}
        &\max \limits_{z \in \Sigma_{K}} \min \limits_{h \in H_{G_P^{\sigma_P}}(w, r)} \langle z - w, h \rangle - 2 E \epsilon  \notag \\
        &\leq \min \limits_{h \in H_{G_P^{\sigma_P}}(w, r)} \langle z^* - w, h \rangle
    \end{align}
    for all $(w, r) \in \mr{int}(\Sigma_K) \times (0, 1)$.
\end{lemma}

Define the curvature of a concave differentiable function $f \colon \mc{A} \to \mathbb{R}$ with respect to a compact set $\mc{A}$ as
\begin{align}
    C_f(\mc{A}) \triangleq \sup \limits_{\substack{x, z \in \mc{A} \\ \alpha \in (0, 1] \\ y = x + \alpha (z -x)}} \frac{1}{\alpha^2} \left( f(x) - f(y) + \langle y - x, \nabla f(x) \rangle \right). \label{eq:Cfdef}
\end{align}
It is known that the curvature constant $C_f(\mc{A})$ controls the error of the \textsf{FW}-type methods for maximizing $f$ over the set~$\mc{A}$~\cite{jaggi2013}. Below, we bound the curvature constant $C_{g_P^{\sigma'}}(\Sigma_{K}^{\gamma})$.

\begin{lemma} \label{lem:curvature}
Let $P \in \Lambda$ and $\sigma' \neq \sigma_P$. Then, for all $\gamma \in (0, \frac{1}{K})$,
    \begin{align}
        C_{g_P^{\sigma'}}(\Sigma_{K}^{\gamma}) \leq \frac{D}{\gamma}, \label{eq:CDgamma}
    \end{align}
    where $D$ is given in \eqref{eq:D}.
\end{lemma}
\begin{IEEEproof}
    In \cite[Lemma 7]{jaggi2013}, Jaggi shows that for a convex (or concave) differentiable function $f$ such that $\nabla f$ is $L$-Lipschitz with respect to a norm $\norm{\cdot}$ over the domain $\mc{D}$, 
    \begin{align}
        C_f(\mc{D}) \leq \mr{diam}_{\norm{\cdot}}(\mc{D})^2 L,
    \end{align}
    where $\mr{diam}_{\norm{\cdot}}(\mc{D})$ denotes the diameter of the set $\mc{D}$ with respect to $\norm{\cdot}$.
    From \lemref{lem:gradLip}, $g_P^{\sigma'}$ is $\frac{D}{\gamma}$-Lipschitz on $\Sigma_K^{\gamma}$. Noting that $\mr{diam}_{\norm{\cdot}_{\infty}}(\Sigma_{K}^{\gamma}) \leq 1$, \lemref{lem:curvature} follows.
\end{IEEEproof}

The following result is used to bound the optimality gap in the $t$-th round of the \textsf{FWS} algorithm. 
\begin{lemma}[\kern-1ex{\cite[Corollary 1]{wang2021fast}}] \label{lem:GPoracle}
    Fix $P \in \Lambda$, $\gamma \in (0, \frac{1}{K})$, $r \in (0, 1)$, $x \in \Sigma_{K}^{\gamma}$, $z \in \Sigma_{K}$, and $\alpha \in (0, \min\{\frac{1}{2}, \frac{r}{L}\})$. Let $y = x + \alpha (z - x)$. Then,
    \begin{align}
        G_P(y) \geq G_P(x) + \alpha \min_{h \in H_{G_P}(x, r)} \langle z -x, h \rangle - \frac{8 D \alpha^2}{\gamma}.
    \end{align}
\end{lemma}
\begin{IEEEproof}
    This result is proved in \cite[Corollary 1]{wang2021fast} for an arbitrary PE problem for which the corresponding score function $g_P^{\sigma'}(w)$ is $L$-Lipschitz in $w$, and its curvature constant satisfies~\eqref{eq:CDgamma}.
\end{IEEEproof}

The following lemma bounds $\Delta_t$ in terms of $\Delta_{t-1}$, $r_t$, and $\epsilon_t$ under a suitably chosen high probability event on the sequence $\{(\tilde{z}(t), \tilde{x}(t))\}_{t \geq 1}$.

\begin{lemma}[{Adapted from \cite[Th.~6]{wang2021fast}}] \label{lem:maxminz}
    Let the sequence $\{r_t\}_{t \geq 1}$ satisfy $L < r_t t$ for all $t \geq 1$, where $L$ is given in \eqref{eq:L}. Let $\epsilon_t > 0$ for all $t \geq 1$, and let $\tilde{z}(t)$ in \eqref{eq:ztilde} and $\tilde{x}(t)$ in \eqref{eq:xtildedef} satisfy
    \begin{align}
        &\max \limits_{z \in \Sigma_{K}} \min \limits_{h \in H_{G_P}(\tilde{x}(t-1), r_t)} \langle z - \tilde{x}(t-1), h \rangle - \epsilon_t \notag \\
        &<  \min \limits_{h \in H_{G_P}(\tilde{x}(t-1), r_t)} \langle \tilde{z}(t) - \tilde{x}(t-1), h \rangle
    \end{align}
    for all $t \notin \mc{I}_{\mr{f}} \cup [K]$.
    Then,
    \begin{align}
        \Delta_t \leq \left(1 - \frac{1}{t}\right) \Delta_{t-1} + \frac{r_t + \epsilon_t}{t}  +  \frac{16 D K}{t^{3/2} \log t}. \label{eq:Deltat}
    \end{align}
    for all $t \notin \mc{I}_{\mr{f}} \cup [K]$.
\end{lemma}
\begin{IEEEproof}
See \appref{app:maxminz}.

\end{IEEEproof}

The next result bounds the optimality gap $\Delta_{T_2}$ under a high probability event on $\{\hat{P}(t)\}_{t \in [T_1, T_2]}$, where $T_1$ and $T_2$ are arbitrary integers.
\begin{lemma}\label{lem:T2gap}
    Let $\epsilon_t > 0$, $L < r_t t$ for all $t \geq 1$,  and let $T_1 < T_2$ be integers such that $T_1 \geq K + 1$ and
     \begin{align}
        &\max \limits_{z \in \Sigma_{K}} \min \limits_{h \in H_{G_P}(\tilde{x}(t-1), r_t)} \langle z - \tilde{x}(t-1), h \rangle - \epsilon_t \notag \\
        &<  \min \limits_{h \in H_{G_P}(\tilde{x}(t-1), r_t)} \langle \tilde{z}(t) - \tilde{x}(t-1), h \rangle \label{eq:highprobevent}
    \end{align}
    holds for $t \in \{T_1, \dots, T_2\} \cap \mc{I}_\mr{f}^{\mr{c}}$. Then,
    \begin{align}
        \Delta_{T_2} &\leq \frac{T_1}{T_2} L + 2L T_2^{-1/2} \log T_2 + \frac{1}{T_2}\sum_{t = 1}^{T_2} \left(r_t + \epsilon_t\right) \notag \\
        &\quad + 32 D K T_2^{-1/2} + \frac{4L}{T_2}.
    \end{align}
\end{lemma}
\begin{IEEEproof}
    See \appref{app:T2gap}.
\end{IEEEproof}

The following corollary to \lemref{lem:T2gap} bounds $\tilde{\Delta}_{T_2}$ under the event in \eqref{eq:highprobevent}.
    
\begin{corollary} \label{cor:GP}
    Under the conditions of \lemref{lem:T2gap}, it holds that
    \begin{align}
        \tilde{\Delta}_{T_2} &\leq \frac{T_1}{T_2} L + 2L T_2^{-1/2} \log T_2 + \frac{1}{T_2}\sum_{t = 1}^{T_2} \left(r_t + \epsilon_t\right) \notag \\
        &\quad + 32 D K T_2^{-1/2} + \frac{L(K + 3)}{T_2}. 
    \end{align}
\end{corollary}
\begin{IEEEproof}
    By \lemref{lem:ctrack}, we have
    \begin{align}
        \norm{w(T_2)  - \tilde{x}(T_2)}_{\infty} \leq \frac{K-1}{T_2}. \label{eq:wT2}
    \end{align}
    Combining \lemref{lem:T2gap} with the $L$-Lipschitzness of $G_P(\cdot)$ (shown in Lemma~\ref{lem:gradg}) and \eqref{eq:wT2}, the corollary follows.
\end{IEEEproof}

\subsection{$\delta$-correctness} \label{sec:deltacorrect}
\lemref{lem:concent}, below, is used to bound the probability that the empirical score function $g_{\hat{P}(t)}^{\sigma_P}(w)$ in \eqref{eq:gsV} evaluated at the true hypothesis $\sigma_P$ exceeds a threshold $\beta$. This result generalizes \cite[Lemma~5]{haghifam}, which proves the case with $M = 1$ equality class and $|\mc{A}_{1}| = 2$, i.e., there are 2 elements in that single cluster.
\begin{lemma} \label{lem:concent}
    Consider $K$ sequences each having lengths $n_i$, $i \in [K]$, where these $K$ sequences are divided into $M$ disjoint clusters, and the sequences within each group $m \in [M]$ are drawn i.i.d. from the same distribution $Q_m \in \mc{P}(\mc{X})$. Let the indices of sequences in group $m$ be denoted by $\mc{A}_m \subseteq [K]$. The sequence with index $i_m \in [K]$ is denoted by $X_{i_m}^{n_{i_m}}$ and follows distribution $P_{i_m}^{n_{i_m}} = Q_{m}^{n_{i_m}}$, where $m$ is the unique group with $i_m \in \mc{A}_{m}$. Let $\hat{P}_i \triangleq \hat{P}_{X_i^{n_i}}$ denote the empirical distribution for the sequence $X_{i}^{n_i}$ and let $w_i = \frac{n_i}{N}$, where $N = \sum_{i \in [K]} n_i$ is the total length of $K$ sequences. Fix $\beta > 0$. Then,  
    \begin{align}
        \Prob{N \sum_{m = 1}^{M} G(\hat{P}_{\mc{A}_m}, w_{\mc{A}_m}) \geq \beta} \leq (N + 1)^{M |\mc{X}|} \exp\{-\beta\}. 
    \end{align}
\end{lemma}
\begin{IEEEproof}
See \appref{app:concent}.
\end{IEEEproof}

We first show that our algorithm has a finite stopping time almost surely.

\subsubsection{Finiteness of $\tau$} Let $P$ be the problem instance, and let $\sigma_P$ be the associated true hypothesis. Fix any alternative hypothesis $\sigma' \neq \sigma_P$. Then, there exists some $m \in [M]$ such that the cluster $\mc{A}_m^{\sigma'}$ has two arms $i_m, j_m \in \mc{A}_m^{\sigma'}$ with distinct distributions $P_{i_m} \neq P_{j_m}$. For $w \in \mr{int}(\Sigma_K)$, define
\begin{align}
    W_m(w, P, \sigma') \triangleq \frac{\sum_{i_m \in \mc{A}_m^{\sigma'}} w_{i_m} P_{i_m}}{\sum_{i_m \in \mc{A}_m^{\sigma'}} w_{i_m}}.
\end{align}

Let $\hat{P}(t) = (\hat{P}_1(t), \dots, \hat{P}_K(t))$ be the empirical distribution of all arms at time $t$, and let $w(t) = (w_1(t), \dots, w_K(t))$ be the fraction of arm pulls at time $t$. By \corref{cor:wtlower}, $w_i(t) \geq \frac{\log t}{K \sqrt{t}} - \frac{K - 1}{t}$ for all $i \in [K]$. Therefore, by the law of large numbers, $\hat{P}_i(t) \to P_i$ as $t \to \infty$.
We would like to show that
\begin{align}
    \liminf_{t \to \infty} \max_{k \in \{i_m, j_m\}} D(\hat{P}_{k}(t) \| W_m(w(t), \hat{P}(t), \sigma')) > 0 \label{eq:liminfD}
\end{align}
holds. First, to eliminate $w(t)$ from the expression, we lower bound the left-hand side of \eqref{eq:liminfD} by
\begin{align}
     \liminf_{t \to \infty} \min_{W \in \mc{P}([K])} \max_{k \in \{i_m, j_m\}} D(\hat{P}_{k}(t) \| W). \label{eq:liminfmin}
\end{align}
For every $\hat{P}_{i_m}(t)$ and $\hat{P}_{j_m}(t)$, the minimum in \eqref{eq:liminfmin} is achieved at 
\begin{align}
    W^*(t) = \alpha(t) \hat{P}_{i_m}(t) + (1-\alpha(t)) \hat{P}_{j_m}(t),
\end{align}
where $\alpha(t) \in [0, 1]$ is such that
\begin{align}
    D(\hat{P}_{i_m}(t) \| W^*(t)) = D(\hat{P}_{j_m}(t) \| W^*(t)). 
\end{align}
Since by assumption, $\min_{k \in [K]} \min_{a \in \mc{X}} P_k(a) \geq p_{\min} > 0$, $W^*(t) \to W^*$ as $t \to \infty$, where
\begin{align}
     W_\alpha^* = \alpha {P}_{i_m} + (1-\alpha) {P}_{j_m},
\end{align}
and $\alpha \in [0, 1]$ is the unique solution to
\begin{align}
    D(P_{i_m} \| W_\alpha^*) = D(P_{j_m} \| W_\alpha^*). \label{eq:DimDjm}
\end{align}
In general, at least one of $D(P_{i_m} \| W_\alpha^*)$ and $D(P_{j_m} \| W_\alpha^*)$ is strictly greater than zero, which follows since $D(P \| Q) = 0$ if and only if $P = Q$. Then, together with \eqref{eq:DimDjm}, this implies that 
\begin{align}
    D(P_{i_m} \| W_\alpha^*) > 0.
\end{align}
Using the fact that $p_{\min} > 0$, we apply the dominated convergence theorem to \eqref{eq:liminfmin} and get
\begin{align}
    \lim_{t \to \infty} \min_{W \in \mc{P}([K])} \max_{k \in \{i_m, j_m\}} D(\hat{P}_{k}(t) \| W) = D(P_{i_m} \| W_\alpha^*) > 0,
\end{align}
which then implies the desired inequality in \eqref{eq:liminfD} for any sequence  $w(t)$. 
From \eqref{eq:liminfD} and using the fact that $w_i(t) \geq \frac{\log t}{K \sqrt{t}} - \frac{K - 1}{t}$, we get
\begin{align}
        \liminf_{t \to \infty} \frac{\max \limits_{k \in \{i_m, j_m\}} t w_{k}(t) D(\hat{P}_{k}(t) \| W_m(w(t), \hat{P}(t), \sigma'))}{\sqrt{t} \log t } > 0. \label{eq:secondliminf}
\end{align}
Therefore, from the definition of $g_{P}^{\sigma'}(w)$ in \eqref{eq:gsV}, for all $\sigma' \neq \sigma_P$, we have
\begin{align}
     \liminf_{t \to \infty} \frac{t g_{\hat{P}(t)}^{\sigma'}(w(t))}{ \sqrt{t} \log t} > 0, \label{eq:liminfg}
\end{align}
which implies 
\begin{align}
     \liminf_{t \to \infty} \frac{Z(t)}{\sqrt{t} \log t } > 0. \label{eq:liminfZ}
\end{align}

Recall from the definition of $\beta(t, \delta)$ in \eqref{eq:betadef} that $\beta(t, \delta)$ scales logarithmically in $t$, hence
\begin{align}
    \lim_{t \to \infty} \frac{\beta(t, \delta)}{\sqrt{t} \log t } = 0.  \label{eq:liminfb}
\end{align}
Combining \eqref{eq:liminfZ} and \eqref{eq:liminfb}, $Z(t)$ exceeds $\beta(t, \delta)$ for some $t \in \mathbb{N}$ almost surely, which implies $\mathbb{P}[\tau < \infty] = 1$.

\subsubsection{Error probability bound} \label{sec:correctness}
In the following, we prove that our algorithm identifies the correct hypothesis $\sigma_P$ with probability at least $1-\delta$. 

The algorithm stops at time
\begin{align}
    \tau \triangleq \inf\{t \geq 1 \colon Z(t) \geq \beta(t, \delta)\}, 
\end{align}
where
\begin{align}
    Z(t) = t \, \min_{\sigma \neq \hat{\sigma}(t)} g_{\hat{P}(t)}^{\sigma}(w(t)) \\
    \hat{\sigma}(t) = \argmin_{\sigma \in \mc{C}} g_{\hat{P}(t)}^{\sigma}(w(t)).
\end{align}
Hence, if the estimate at the stopping time $\tau$ is erroneous, then the score $\tau g_{\hat{P}(\tau)}^{\sigma_P}(w(\tau))$ corresponding to the true hypothesis exceeds the threshold $\beta(\tau, \delta)$, i.e., 
we have $\{ \hat{\sigma}(\tau) \neq \sigma_P\} \subseteq \{\tau g_{\hat{P}(\tau)}^{\sigma_P}(w(\tau)) \geq \beta(\tau, \delta)\}$.

Consider $K$ infinite-length sequences $\tilde{X}_i^{\infty} = (\tilde{X}_{i, 1}, \tilde{X}_{i, 2}, \dots)$, $i \in [K]$, where $\tilde{X}_{i, n}$ represents the outcome for arm $i$ at its $n$-th pull. In other words, $(\tilde{X}_i^{\infty} \colon i \in [K])$ are the stacked outcome values that are revealed to the decision maker according to the order decided by the algorithm. The random variables $(\tilde{X}_{i, n})_{i \in [K], n \in \mathbb{N}}$ are jointly independent and $\tilde{X}_{i, n} \sim P_i$. Let $\tilde{P}_{i}(t_i)$ denote the empirical distribution $\hat{P}_{\tilde{X}_i^{t_i}}$ for $i \in [K]$. Let $\tilde{P}(t_{[K]}) = (\tilde{P}_{i}(t_i) \colon i \in [K])$ and $\tilde{w}(t_{[K]}) = \frac{1}{\sum_{i \in [K]} t_i}(t_1, \dots, t_K)$. Note that for $t_i = t w_i(t)$ for all $i \in [K]$, we have $\hat{P}(t) = \tilde{P}(t_1, \dots, t_K)$ and $w(t) = \tilde{w}(t_1, \dots, t_K)$.

Let $[B]$ be the indices of arms that are in the unconstrained group, i.e., $\mc{A}_{M + 1}^{\sigma_P} = [B]$ with $B < K$. Consider two allocation vectors $t_{[K]}^{(1)}$ and $t_{[K]}^{(2)}$, where $t_i^{(j)}$ is a non-negative integer for all $i \in [K]$ and $j \in [2]$, and $\sum_{i = 1}^K t_i^{(j)} = t$ for $j \in [2]$. Let $t_{[B+1:K]}^{(1)} = t_{[B + 1:K]}^{(2)}$, $\sum_{i \in [B]} t_i^{(1)} = t^* < t$, and $t_{[B]}^{(2)} = (t^*, 0, \dots, 0)$. By the definition of $g_Q^{\sigma_P}(w)$ in \eqref{eq:gsV}, $g_Q^{\sigma_P}(w)$ is independent of $Q_{[B]}$ and $w_{[B]}$. Therefore, from the above argument, it follows that $g_{\tilde{P}(t_{[K]}^{(1)})}(\tilde{w}(t_{[K]}^{(1)})) = g_{\tilde{P}(t_{[K]}^{(2)})}(\tilde{w}(t_{[K]}^{(2)}))$.

Recall that 
\begin{align}
    \tilde{K} = \max_{\sigma \in \mc{C}} \left \lvert \bigcup_{m \in [M]} \mc{A}_m^{\sigma} \right \rvert
\end{align}
is the maximum number of arms that belong to a cluster. 
Using the arguments above, we define
\begin{align}
    \mc{T}(t) &\triangleq \bigg\{(t_1, \dots, t_K) \colon t_i \in \mathbb{N} \, \forall \, i \in [K], \sum_{i \in \bigcup \limits_{m \in [M]} \mc{A}_m^{\sigma_P}} t_i \leq t, \notag \\
    &t_j = t - \sum_{i \in \bigcup \limits_{m \in [M]} \mc{A}_m^{\sigma_P}} t_i \text{ where } j = \min_{j' \in [K] \setminus \bigcup \limits_{m \in [M]} \mc{A}_m^{\sigma_P}} j' \bigg\}.
\end{align}
The set $\mc{T}(t)$ refers to the all possible allocations $(t_1, \dots, t_K)$ of $t$ pulls, where only one arm (with index $j$) is pulled from the unconstrained group $[K] \setminus \bigcup \limits_{m \in [M]} \mc{A}_m^{\sigma_P}$.

From the construction of the bandit algorithm, there exists some $t_{[K]} \in \mc{T}(t)$ such that
\begin{align}
    g_{\hat{P}(t)}^{\sigma_P}(w(t)) = g_{\tilde{P}(t_{[K]})}^{\sigma_P}(\tilde{w}(t_{[K]})),
\end{align}
which implies that for any $\beta \in \mathbb{R}$,
\begin{align}
    \{ g_{\hat{P}(t)}^{\sigma_P}(w(t)) \geq \beta \}  \subseteq \bigcup_{t_{[K]} \in \mc{T}(t)} \{ g_{\tilde{P}(t_{[K]})}^{\sigma_P}(\tilde{w}(t_{[K]})) \geq \beta\}. \label{eq:gunion}
\end{align}

We bound the error probability of our algorithm as
\begin{align}
    &\Prob{\hat{\sigma}(\tau) \neq \sigma_P} \notag \\
    &\quad \leq \Prob{\tau g_{\hat{P}(\tau)}^{\sigma_P}(w(\tau)) \geq \beta(\tau, \delta)} \\
   &\quad\leq \Prob{\bigcup_{t = 1}^{\infty} \{t g_{\hat{P}(t)}^{\sigma_P}(w(t)) \geq \beta(t, \delta)\}} \\
   &\quad\leq \sum_{t = 1}^{\infty}  \Prob{t g_{\hat{P}(t)}^{\sigma_P}(w(t)) \geq \beta(t, \delta)} \label{eq:union1}\\
   &\quad\leq  \sum_{t = 1}^{\infty} \sum_{t_{[K]} \in \mc{T}(t)} \Prob{t g_{\tilde{P}(t_{[K]})}^{\sigma_P}(\tilde{w}(t_{[K]})) \geq \beta(t, \delta)} \label{eq:Xtildecons}\\
   &\quad \leq \sum_{t = 1}^{\infty} (t + 1)^{\tilde{K}} (t + 1)^{M |\mc{X}|} \exp\{-\beta(t, \delta)\} \label{eq:concentused}\\
   &\quad \leq \delta. \label{eq:sumt2}
\end{align}
Here, \eqref{eq:union1} follows from the union bound; \eqref{eq:Xtildecons} follows from \eqref{eq:gunion} and the union bound; \eqref{eq:concentused} applies \lemref{lem:concent} and uses the fact that $|\mc{T}(t)| \leq (t+1)^{\tilde{K}}$; and \eqref{eq:sumt2} follows from the definition of $\beta(t, \delta)$ in \eqref{eq:betadef} and that $\sum_{t = 2}^{\infty} \frac{1}{t^2} = \frac{\pi^2}{6} - 1$.

\subsection{Expected Value of the Stopping Time}
\subsubsection{Establishing a sufficient condition for $\hat{\sigma}(t) = \sigma_P$}
Let 
\begin{align}
    r_t = t^{-b_0}, \label{eq:rtb0}
\end{align}
where $b_0 \in (0, 1)$ is going to be determined later. To satisfy the condition $L < r_t t$, we assume that 
\begin{align}
    t \geq L^{\frac{1}{1 - b_0}} + 1.
\end{align}

\label{sec:suff} Suppose that $\lVert \hat{P}(t) - P \rVert_{\infty} \leq \epsilon_t$ for $t \geq 1$. We here establish a condition for $\epsilon_t$ that suffices to have $\hat{\sigma}(t) = \sigma_P$. According to \eqref{eq:sigmahat}, $\hat{\sigma}(t) = \sigma_P$ if
\begin{align}
    \min_{\sigma' \neq \sigma_P} g_{\hat{P}(t)}^{\sigma'}(w(t)) > g_{\hat{P}(t)}^{\sigma_P}(w(t)). \label{eq:condition}
\end{align}
By Assumption \ref{as:finer}, every alternative hypothesis $\sigma' \neq \sigma_P$ has at least one cluster $m \in [M]$ such that there exist arms $i \neq j \in \mc{A}_m^{\sigma'}$ with $P_{i} \neq P_{j}$. Further 
define
\begin{align}
    d_{\min} \triangleq \min_{\sigma' \neq \sigma_P} \max_{\substack{m \in [M] \\ i \neq j \in \mc{A}_m^{\sigma'}} } \norm{P_i - P_j}_{\infty} \label{eq:dmin}
\end{align}
and assume that
\begin{align}
    \epsilon_t \leq \frac{d_{\min}}{2}. 
\end{align}
Define the constant
\begin{align}
    D_1 \triangleq \min_{\substack{\sigma' \neq \sigma_P \\ W \in \mc{P}(\mc{X}) \\ \tilde{P}_i \colon \norm{P_i - \tilde{P}_i}_{\infty} \leq \frac{d_{\min}}{3} \\ \tilde{P}_j \colon \norm{P_j - \tilde{P}_j}_{\infty} \leq \frac{d_{\min}}{3}}} \max_{\substack{m \in [M] \\ i \neq j \in \mc{A}_m^{\sigma'}} } \max\{ D(\tilde{P}_i \| W), D(\tilde{P}_j \| W)\}. \label{eq:D1}
\end{align}
By \eqref{eq:dmin}, $\tilde{P}_i$ and $\tilde{P}_j$ in \eqref{eq:D1} are distinct and since $D(P_a \| P_b) = 0$ if and only if $P_a = P_b$, we have $D_1 > 0$. 

By Corollary~\ref{cor:wtlower}, for $t \geq 4 K^4$, 
\begin{align}
    w_i(t) > \frac{t^{-1/2} \log t}{2K} \label{eq:wlower}
\end{align}
for all $i \in [K]$. Combining \eqref{eq:D1} and \eqref{eq:wlower} with the definition of $g_P^{\sigma}(w)$ in \eqref{eq:gsV} and \lemref{lem:inf}, we have
\begin{align}
    \min_{\sigma' \neq \sigma_P} g_{\hat{P}(t)}^{\sigma'}(w(t)) > \frac{t^{-1/2} \log t}{2K} D_1.
\end{align}

Assume that $\epsilon_t$ satisfies
\begin{align}
    \epsilon_t \leq \frac{p_{\min}}{2}, \label{eq:etpmin}
\end{align}
where $p_{\min}$ is defined in \eqref{eq:pmin}.
Using the definition of $G(P_{\mc{A}}, w_{\mc{A}})$ in \eqref{eq:GA} and $g_{P}^{\sigma}(w)$ in \eqref{eq:gsV}, given $\lVert \hat{P}(t) - P \rVert_{\infty} \leq \epsilon_t$, we have
\begin{align}
    g_{\hat{P}(t)}^{\sigma_P}(w(t)) \leq \max_{i \in [K]} \max_{\substack{ \tilde{P}_i \colon \norm{\tilde{P}_i - P_i}_{\infty} \leq \epsilon_t \\ \tilde{W}_i \colon \norm{\tilde{W}_i - P_i}_{\infty} \leq \epsilon_t}} D(\tilde{P}_i \| \tilde{W}_i). \label{eq:gpupper}
\end{align}
Applying the reverse Pinsker's inequality from \cite[Sec.~7.6]{polyanskiyLectureNotes} to the right hand side of \eqref{eq:gpupper}, we get
\begin{align}
    g_{\hat{P}(t)}^{\sigma_P}(w(t)) &\leq \max_{i \in [K]} \frac{1}{2 \min_{a \in \mc{X}} \tilde{W}_i(a)} \norm{\tilde{P}_i - \tilde{W}_i}_1^2 \\
    &\leq \max_{i \in [K]} \frac{K^2}{p_{\min}} \norm{\tilde{P}_i - \tilde{W}_i}_\infty^2 \\
    &\leq \frac{4 K^2 \epsilon_t^2}{p_{\min}}.
\end{align}

Hence, we set
\begin{align}
    F &\triangleq \sqrt{\frac{D_1 p_{\min}}{8 K^3}} \label{eq:Fdef}\\
    \epsilon_t &= F t^{-1/4} \sqrt{\log t}, \label{eq:et}
\end{align}
which guarantees 
\eqref{eq:condition} on the event $\big\{\lVert \hat{P}(t) - P \rVert_{\infty} \leq \epsilon_t\big\}$ for 
\begin{align}
    t \geq t_0 \triangleq \max\left\{L^{\frac{1}{1 - b_0}} + 1, 4 K^4, \left( \frac{D_1 p_{\min}}{K^3 \min\{d_{\min}^2, p_{\min}^2\}}\right)^4 \right\}. \label{eq:tmax}
\end{align}
In \eqref{eq:tmax}, we use the fact that $\frac{t}{\log^2(t)} \geq \frac{1}{4} \sqrt{t}$ for $t > 1$ and that
\begin{align}
    \epsilon_t \leq \min \left\{\frac{p_{\min}}{2}, \frac{d_{\min}}{2} \right\} \iff \frac{t}{\log^2(t)} \geq \frac{F^4}{(\min\{\frac{p_{\min}}{2}, \frac{d_{\min}}{2}\})^4}. 
\end{align}

As opposed to \cite{garivier2016, wang2021fast, prabhu2022}, we choose $\epsilon_t$ in \eqref{eq:et} as a decaying function of $t$ rather than as an arbitrary but constant $\epsilon$; our choice in \eqref{eq:et} is important to optimize the second-order term in \thmref{thm:upper} in terms of $\log \frac{1}{\delta}$. 

\begin{remark}
    We choose the forced exploration steps in \eqref{eq:If} for which $w_i(t) = \Omega(t^{-1/2} \log t)$ differently than \cite{garivier2016, wang2021fast} because \eqref{eq:If} allows us to set the value of $\epsilon_t$ as in \eqref{eq:et}. The choice of $\epsilon_t$ in \eqref{eq:et} simultaneously guarantees the condition in \eqref{eq:condition} and that the term $\sum_{T = \max\{T_0(\delta), t_0\}}^{\infty} \Prob{\mc{E}(T)^{\mr{c}}}$ in \eqref{eq:ETbound}, below, decays to zero as $\delta \to 0$. With the forced exploration steps in \cite{garivier2016, wang2021fast} for which only $w_i(t) = \Omega(t^{-1/2})$ holds, no such $\epsilon_t$ sequence can be found to satisfy both conditions.
\end{remark}

\subsubsection{High probability events}
Let $0 < b_1 < b_2 < 1$ be some constants that are going to be optimized later. Define the functions 
\begin{align}
    h_1(T) \triangleq \lceil T^{b_1} \rceil, \quad h_2(T) \triangleq \lceil T^{b_2} \rceil 
\end{align}
for $T \geq 1$ and the high probability events
\begin{align}
    \mc{E}_{1,t} &\triangleq \bigg\{ \max \limits_{z \in \Sigma_{K}} \min \limits_{h \in H_{G_P^{\sigma_P}}(\tilde{x}(t-1), r_t)} \langle z - \tilde{x}(t-1), h \rangle - 2 E \epsilon_t  \notag \\*
        &< \min \limits_{h \in H_{G_P^{\sigma_P}}(\tilde{x}(t-1), r_t)} \langle \tilde{z}(t) - \tilde{x}(t-1), h \rangle \bigg\} \label{eq:E1t}
\end{align}
for $t \in \mc{I}_{\mr{f}}^{\mr{c}}$ and $\Prob{\mc{E}_{1, t}} = 1$ if $t \in \mc{I}_{\mr{f}}$, and
\begin{align}
   \mc{E}_{2,t} &\triangleq \{ \hat{\sigma}(t) = \sigma_P \} \bigcap \{ G_{\hat{P}(t)}^{\sigma_P}(w(t)) \geq  G_{P}^{\sigma_P}(w(t)) - E \epsilon_t\} \label{eq:E2t}\\
   \mc{E}(T) &\triangleq \bigcap_{t = h_1(T)}^{T} (\mc{E}_{1,t} \cap \mc{E}_{2,t}). \label{eq:ETdef}
\end{align}
From \lemref{lem:LipinP}, \lemref{lem:LipinFW}, and the above arguments in \secref{sec:suff}, 
\begin{align}
    \mc{E}(T) \supseteq \bigcap_{t = h_1(T)}^{T} \left\{\norm{\hat{P}(t) - P}_{\infty} < \epsilon_t \right\}. \label{eq:ETcap}
\end{align}

\subsubsection{Bounding the expected value of the stopping time}
The following result is a concentration inequality for the $\ell_{\infty}$-norm of the gap between the empirical distribution and the true distribution.
\begin{lemma}\label{lem:conclinf}
    Let $X_1, \dots, X_n$ be i.i.d. following a distribution $P$ on a finite alphabet $\mc{X}$. Let $\hat{P}_n$ be the empirical distribution associated with $X^{n} = (X_1, \dots, X_n)$. Then, for any $\epsilon > 0$,
    \begin{align}
        &\Prob{\norm{\hat{P}_n - P}_{\infty} \geq \epsilon} \notag \\
        &\leq 2 C_1  (|\mc{X}|-1)(C_0 n)^{\frac{|\mc{X}|}{2}-1}  \exp\{- 2 n \epsilon^2\},
    \end{align}
    where $C_0 \approx 3.1967$ and $C_1 \approx 2.9290$ are constants.
\end{lemma}
\begin{IEEEproof}
Let $\mr{TV}(P, Q) \triangleq \sup_{\mc{A} \subseteq \mc{X}} |P(\mc{A}) - Q(\mc{A})|$ denote the total variation distance between $P$ and $Q$. By Pinsker's inequality (e.g., \cite[Th.~7.10]{polyanskiyLectureNotes}), 
\begin{align}
    D(P \| Q) \geq 2 (\mr{TV}(P, Q))^2 \geq 2 \norm{P - Q}_{\infty}^2.
\end{align}
Hence,
\begin{align}
    &\Prob{\big \lVert \hat{P}_n - P \big \rVert_{\infty} \geq \epsilon} \notag \\
    &\leq \Prob{D(\hat{P}_n \| P) \geq 2 \epsilon^2} \\
    &\leq  C_1  \left((|\mc{X}|-2)(C_0 n)^{\frac{|\mc{X}|}{2}-1} + 1\right) \exp\{- 2 n \epsilon^2\} \label{eq:mardia} \\
    &\leq 2 C_1  (|\mc{X}|-1)(C_0 n)^{\frac{|\mc{X}|}{2}-1}  \exp\{- 2 n \epsilon^2\},
\end{align}
where \eqref{eq:mardia} follows from the concentration bound in \cite[Th.~3]{mardia2020conc}.
\end{IEEEproof}

\begin{lemma} \label{lem:ETlemma}
It holds that
    \begin{align}
        \lim_{\delta \to 0} \sum_{T = \max\{T_0(\delta), t_0\}}^{\infty} \Prob{\mc{E}(T)^{\mr{c}}} = 0,\label{eq:ETsum}
    \end{align}
\end{lemma}
where $T_0(\delta)$ is an arbitrary function of $\delta$ such that $T_0(\delta) \to \infty$ as $\delta \to 0$. We specify its value in \eqref{eq:T0def} below.
\begin{IEEEproof}
    From \eqref{eq:ETcap} and the union bound, we have
    \begin{align}
        \Prob{\mc{E}(T)^{\mr{c}}} &\leq \sum_{t = h_1(T)}^{T} \Prob{\norm{\hat{P}(t) - P}_{\infty} \geq \epsilon_t} \\
        &\leq \sum_{t = h_1(T)}^{T} \sum_{i = 1}^K \Prob{\norm{\hat{P}_i(t) - P_i}_{\infty} \geq \epsilon_t}. \label{eq:ETC}
    \end{align}
    Similar to the proof in \secref{sec:correctness}, let $\tilde{X}_{i, n}$ be the outcome from the $i$-th arm at the $n$-th pull and let $\tilde{P}_i(s)$ be the empirical distribution of the sequence $(\tilde{X}_{i, n})_{n = 1}^{s}$.

    Define the constants
    \begin{align}
        B_1 &\triangleq \frac{K C_1  (|\mc{X}|-1)C_0^{\frac{|\mc{X}|}{2}-\frac{1}{2}}}{F^2 (1 - \frac{p_{\min}^2}{2})} \label{eq:B1}\\
        B_2 &\triangleq \frac{F^2 b_1^2}{K} \label{eq:B2} \\
        B_3 &\triangleq \frac{|\mc{X}|}{2} + \frac{1}{2}. \label{eq:B3}
    \end{align}
    
    For $t \geq t_0$, from \eqref{eq:wlower}, we have $N_i(t) \geq \lceil \frac{t^{1/2}\log t}{2K} \rceil$. Therefore, by the union bound, we get
    \begin{align}
        &\Prob{\norm{\hat{P}_i(t) - P_i}_{\infty} \geq \epsilon_t}\notag\\
        &\leq \sum_{s = \lceil \frac{t^{1/2}\log t}{2K} \rceil}^{t} \Prob{\norm{\tilde{P}_i(s) - P_i}_{\infty} \geq \epsilon_t} \\
        &\leq  2 C_1  (|\mc{X}|-1)(C_0 t)^{\frac{|\mc{X}|}{2}-1} \sum_{s = \lceil \frac{t^{1/2}\log t}{2K} \rceil}^{t} \exp\{- 2 s \epsilon_t^2\} \label{eq:mardiaused} \\
        &\leq \frac{ 2 C_1  (|\mc{X}|-1)(C_0 t)^{\frac{|\mc{X}|}{2}-1}}{1 - \exp\{-2 \epsilon_t^2\}} \exp\left\{ - \frac{t^{1/2} \log t}{K} \epsilon_t^2 \right\} \label{eq:geosum} \\
        &\leq \frac{t^{1/2}}{\log t} \frac{C_1  (|\mc{X}|-1)(C_0 t)^{\frac{|\mc{X}|}{2}-1}}{F^2 (1 - 2F^2 t^{-1/2} \log t)} \exp\left\{ - \frac{F^2}{K} \log^2 t \right\} \label{eq:etused} \\
        &\leq \frac{B_1}{K} \exp\left\{ - \frac{F^2}{K} \log^2 t + \left(\frac{|\mc{X}|}{2}-\frac{1}{2} \right) \log t \right\}, \label{eq:pminineq}
    \end{align}
    where \eqref{eq:mardiaused} follows from \lemref{lem:conclinf}, \eqref{eq:geosum} applies the geometric sum formula, \eqref{eq:etused} uses the value of $\epsilon_t$ in \eqref{eq:et} and the fact that $\exp\{-x\} \leq 1-x + x^2$ for $x \geq 0$, and \eqref{eq:pminineq} uses \eqref{eq:etpmin}.

    From \eqref{eq:B1}--\eqref{eq:B3}, \eqref{eq:ETC}, \eqref{eq:pminineq}, and the fact that $h_1(T) \geq T^{1/2}$, we get
    \begin{align}
        \Prob{\mc{E}(T)^{\mr{c}}} &\leq B_1 \exp\left\{ - \frac{F^2}{K} \log^2 h_1(T) + B_3 \log T \right\} \\
        & \leq B_1 \exp\left\{ - B_2 \log^2 T + B_3 \log T \right\}.
    \end{align}
    Finally, we bound the sum in \eqref{eq:ETsum} using the integral approximation as
    \begin{align}
        &\sum_{T = \max\{T_0(\delta), t_0\}}^{\infty} \Prob{\mc{E}(T)^{\mr{c}}} \notag \\
        &\leq \int \limits_{T_1(\delta)}^{\infty} \exp\left\{ - B_2 \log^2 T + B_3 \log T \right\} \mr{d} T\\
        &= \frac{B_1}{2 \sqrt{B_2}} e^{\frac{(B_3 + 1)^2}{4 B_2}}\frac{\exp\{-B_2 \log^2 T_1(\delta)\}}{\sqrt{B_2} \log T_1(\delta)} (1 + o(1)) \label{eq:expT1}
    \end{align}
    as $\delta \to 0$, where 
    \begin{align}
        T_1(\delta) \triangleq \max\{T_0(\delta), t_0\} - 1 \to \infty.
    \end{align}
    The expression in \eqref{eq:expT1} approaches 0 as $T_1(\delta)$ approaches infinity, equivalently, as $\delta$ approaches 0.
    
\end{IEEEproof}

In the following, we derive a second-order asymptotic bound on the expected stopping time $\E{\tau}$.

Let $T \in \mathbb{N}$ be an arbitrary integer. Recall from \eqref{eq:rtb0} that $r_t = t^{-b_0}$. Define the function
\begin{align}
    \psi(t) &\triangleq  \frac{11EF}{3} t^{-1/4} \sqrt{\log t} + \frac{1}{1-b_0} t^{-b_0} \notag \\
    &\quad + L t^{- \frac{b_2 - b_1}{b_2}} + 2 L t^{-1/2} \log t + 32 D K t^{-1/2} \notag \\
    &\quad + L(K + 4) t^{-1}. \label{eq:psi}
\end{align}

On the event $\mc{E}(T)$, for $h_1(T) \geq t_0$ where $t_0$ is defined in \eqref{eq:tmax}, we have
\begin{align}
    &\min\{\tau, T\} \notag \\
    &\leq h_2(T) + \sum_{t = h_2(T)}^T 1\{ \tau > t \} \\
    &\leq h_2(T) + \sum_{t = h_2(T)}^T 1\{t G_{\hat{P}(t)}^{\hat{\sigma}(t)}(w(t)) < \beta(t, \delta) \} \label{eq:defstop}\\
    &\leq h_2(T) + \sum_{t = h_2(T)}^T 1\{t (G_{P}^{\sigma_P}(w(t)) - E \epsilon_t) < \beta(t, \delta) \} \label{eq:GPresult} \\
    &\leq T^{b_2} + 1 + \sum_{t = h_2(T)}^T 1\{t (T^*(P) - \psi(t)) < \beta(t, \delta) \} \label{eq:GPresult2}\\
    &\leq T^{b_2} + 1 + \tilde{t}(\delta) \label{eq:sumbeta}
\end{align}
where $\tilde{t}(\delta)$ is defined as the (largest) solution\footnote{For $\delta$ small enough, the solution is unique.} to the equation
\begin{align}
t (T^*(P) - \psi(t)) = \beta(t, \delta). \label{eq:tNewton}
\end{align}
Here, \eqref{eq:defstop} follows from the definition of the stopping time $\tau$, and \eqref{eq:GPresult} follows from \eqref{eq:E2t}--\eqref{eq:ETdef}. The inequality \eqref{eq:GPresult2} applies \corref{cor:GP} with $T_2$ replaced by $t$ and $\epsilon_t$ replaced by $2E \epsilon_t$ and uses the fact that $h_1(T) \geq T^{b_1} + 1$ and $T^{b_2} + 1 \geq h_2(T) \geq T^{b_2}$. See \appref{app:GPresult} for the derivation of \eqref{eq:GPresult2}. The inequality \eqref{eq:sumbeta} holds since $\beta(t, \delta) - t (T^*(P) - \psi(t))$ is decreasing for $t \geq \tilde{t}(\delta)$.

Define the quantity
\begin{align}
    T_0(\delta) \triangleq \inf \left\{ T \in \mathbb{N} \colon T^{b_2} + 1 +   \tilde{t}(\delta) \leq T \right\}. \label{eq:T0def}
\end{align}
Next, we asymptotically solve $T$ from the equation 
\begin{align}
    T^{b_2} + 1 +  \tilde{t}(\delta) = T \label{eq:asympT}
\end{align}
as $\delta \to 0$. Recall that $\beta(t, \delta) = \log \frac{1}{\delta} + O(\log(t))$ and $\psi(T) \to 0$ as $t \to \infty$. Therefore, the first-order approximations to $\tilde{t}(\delta)$ and $T_0(\delta)$ are given by
\begin{align}
    \tilde{t}(\delta) &= \frac{\log\frac{1}{\delta}}{T^*(P)} (1 + o(1)) \\
    T_0(\delta) &= \frac{\log\frac{1}{\delta}}{T^*(P)} (1 + o(1)). \label{eq:T0asym}
\end{align}
We set
\begin{align}
    b_0 = \frac{4}{5}, \quad 
    b_1 = \frac{3}{8}, \quad
    b_2 = \frac{3}{4}. 
\end{align}
Under this choice of parameters, $\psi(t)$ in \eqref{eq:psi} satisfies
\begin{align}
    \psi(t) =  \frac{11EF}{3} t^{-1/4} \sqrt{\log t} + O(t^{-1/2} \log t).
\end{align}
Next, we apply Newton's method to the equation \eqref{eq:tNewton} with the initial estimate of $\tilde{t}_0 = \frac{\log\frac{1}{\delta}}{T^*(P)}$ and get
\begin{align}
    \tilde{t}(\delta) &= \tilde{t}_0 + \frac{11EF}{3} \tilde{t}_0^{3/4} \sqrt{\log \tilde{t}_0} + O(\tilde{t}_0^{1/2} \log \tilde{t}_0) \\
    &= \frac{\log\frac{1}{\delta}}{T^*(P)}  \notag \\
    &\quad \cdot \bigg[ 1 +  \frac{11EF}{3\, (T^*(P))^{3/4}}  \left(\log \frac{1}{\delta}\right)^{-1/4} \sqrt{ \log \log \frac{1}{\delta}} \notag \\
    &\quad \quad + O\bigg(\left(\log \frac{1}{\delta}\right)^{-1/2} \log \log \frac{1}{\delta} \bigg) \bigg].
\end{align}
Similarly, applying Newton's method to the equation \eqref{eq:asympT} gives
\begin{align}
    T_0(\delta) &= \frac{\log\frac{1}{\delta}}{T^*(P)}  \notag \\
    &\quad \cdot \bigg[ 1 +  \frac{11EF}{3\, (T^*(P))^{3/4}}  \left(\log \frac{1}{\delta}\right)^{-1/4} \sqrt{ \log \log \frac{1}{\delta}} \notag \\
    &\quad \quad + O\bigg(\left(\log \frac{1}{\delta}\right)^{-1/4} \bigg) \bigg]. \label{eq:T0asymsec}
\end{align}

From \eqref{eq:sumbeta}--\eqref{eq:T0def}, for $T \geq \max\{T_0(\delta), t_0\}$, we have $\mc{E}(T) \subseteq \{\tau \leq T\}$. Hence, 
\begin{align}
    \E{\tau} &= \sum_{T = 0}^{\infty} \Prob{\tau > T} \\
    &\leq T_0(\delta) + t_0 + \sum_{T = \max\{T_0(\delta), t_0\}}^{\infty} \Prob{\mc{E}(T)^{\mr{c}}}.\label{eq:ETbound}
\end{align}
From \lemref{lem:ETlemma}, \eqref{eq:T0asymsec}, and \eqref{eq:ETbound}, we conclude 
\begin{align}
    &\E{\tau} \leq \frac{\log\frac{1}{\delta}}{T^*(P)} \left( 1 + O\left( \left(\log \frac{1}{\delta} \right)^{-1/4} \sqrt{\log \log \frac{1}{\delta}} \right) \right). 
\end{align}

\section{Discussions} \label{sec:disc}
\subsection{Discussion of the Algorithm} \label{sec:discAlgo}
In this section, we discuss on our algorithm \textsf{TaS-FW} from different aspects. 
\subsubsection{Current best estimate}
Our algorithm \textsf{TaS-FW} estimates the true hypothesis as
\begin{align}
    \hat{\sigma}(t) = \argmin_{\sigma \in \mc{C}} g_{\hat{P}(t)}^{\sigma}(w(t)),
\end{align}
where $g_P^{\sigma}(w)$ is given in \eqref{eq:gsV}. 
In the standard unstructured BAI where the goal is to identify the arm with the largest mean, the current best estimate is simply the arm with the largest empirical mean and is independent of the empirical allocation $w(t)$. This follows because in the unstructured BAI problem, every empirical distribution $\hat{P}(t)$ defines a valid problem instance, ensuring that the maximum log-likelihood without constraints (see the hypothesis $H_1$ in \eqref{eq:hypoH1}) is equal to the maximum log-likelihood under $\sigma_{\hat{P}(t)}$.   
However, in our clustering problem, $\sigma_{\hat{P}(t)}$ is almost always undefined because $\Lambda$ does not include all sets of empirical problem instances observable at time $t$. Since the most likely valid problem instance $Q \in \Lambda$ in the vicinity of $\hat{P}(t)$ depends on the allocation $w(t)$, $\hat{\sigma}(t)$ depends both on the empirical distribution $\hat{P}(t)$ and on the allocation $w(t)$. 

Unlike in \cite{garivier2016, wang2021fast}, the current best estimate $\hat{\sigma}(t)$ depends on the allocation vector $w(t)$ up to time $t$. Specifically, if $w_{\mc{A}_m^{\sigma}}$ has a single non-zero coordinate for all clusters $m \in [M]$, then $g_{\hat{P}(t)}^{\sigma}(w(t)) = 0$, and $\sigma$ achieves the minimum in \eqref{eq:sigmahat} regardless of the value of $\hat{P}(t)$. This makes it impossible to estimate the mapping correctly as $\sigma_P$ when $w(t)$ satisfies the above condition even if $\lVert \hat{P}(t) - P \rVert_{\infty}$ is arbitrarily small. To guarantee that $\hat{\sigma}(t) = \sigma_P$ when $\lVert \hat{P}(t) - P \rVert_{\infty}$ is sufficiently small, through the procedure in \eqref{eq:ztilde}--\eqref{eq:xtildedef}, we track a sequence whose components scale as $\Omega(t^{-1/2} \log t)$, rather than $\Omega(t^{-1/2})$ as in \cite{garivier2016, wang2021fast}.

\subsubsection{Difficulty in Solving the Optimal Allocation and the \textsf{FWS} Algorithm} \label{sec:discdifficult}

Ideally, we would like to compute the optimal allocation
\begin{align}
    w^*(t) &= \argmax_{w \in \Sigma_{K}} G_{\hat{P}(t-1)}^{\hat{\sigma}(t-1)}(w) \label{eq:wstar2} \\
    &= \argmax_{w \in \Sigma_{K}} \min_{\sigma' \in \mc{C} \setminus \{\hat{\sigma}(t-1)'\}} g_{\hat{P}(t-1)}^{\sigma'}(w)
\end{align}
at each time $t$.

Solving \eqref{eq:wstar2} simplifies for some clustering problems. For instance, for the problem of matching pairs from two groups given in Example 1, if the number of nominal arms (the number of clusters) is $M = 1$, then the optimization problem in \eqref{eq:wstar2} reduces to that in \cite[Th.~5]{garivier2016}. Specifically, $P_1$ plays the role of the optimal arm $\mu_1$ in \cite[Th.~5]{garivier2016}, the candidate arm with distribution $P_{\sigma(1)} = P_1$ is dropped from the optimization problem with $w^*_{\sigma(1)}(t)$ set to 0, and the remaining $K-2$ arms following distributions $P_{[K] \setminus \{1, \sigma(1)\}}$ play the role of the sub-optimal arms $\mu_2, \dots, \mu_{K-1}$ in \cite[Th.~5]{garivier2016}. Garivier and Kaufmann \cite{garivier2016} show that the optimal allocation satisfies the relationship
\begin{align}
    G((P_1, P_j), (w_1^*(t), w_j^*(t))) = G((P_1, P_k), (w_1^*(t), w_k^*(t)))
\end{align}
for all $j, k \in [K] \setminus \{1, \sigma(1)\}$. Using this property, they efficiently solve \eqref{eq:wstar2} by a simple procedure based on a line search strategy. A similar line search strategy is possible for the standard odd arm identification with $M = 1$ described in Example 2 in \secref{sec:notation}. 

For PE problems, solving the oracle given by \eqref{eq:3opt} can be difficult in general. For general clustering problems, which we focus on, introducing a minimum distance constraint between two hypotheses such as $\norm{P - Q}_{\infty} \geq \epsilon$ for every $P$ and $Q$ such that $\sigma_P \neq \sigma_Q$ would make the inner infimum problem non-convex, thereby leading to a challenging optimization problem.  Even for the problems where the inner infimum problem in \eqref{eq:3opt} is convex, solving the oracle \eqref{eq:wstar2} can be computationally intensive \cite{agrawalHeavy}. For example, for the matching pairs problem given in Example 1, when the number of clusters $M \geq 2$,  \eqref{eq:gsV} and \eqref{eq:GsV} indicate that $w_{j}$, $j > M$, appears in multiple alternative hypotheses~$\sigma'$, making the simple line search strategy infeasible. To address the issue of solving \eqref{eq:wstar} for general bandit structures where finding the exact solution is not possible or computationally intensive, Wang \emph{et al.} \cite{wang2021fast} develop a computationally-efficient \textsf{FWS} algorithm that only requires to solve a single linear program at each time $t$. The \textsf{FWS} algorithm relies on the assumption that for each hypothesis $\sigma \in \mc{C}$, the set of problem instances $\Lambda_{\sigma}$ is an open set relative to $\Lambda$, and its complement $\Lambda \setminus \Lambda_{\sigma}$ is a finite union of convex sets. This assumption ensures that the inner infimum problem in \eqref{eq:3opt} is a convex program and that it is continuously differentiable on $(\mr{int}(\Sigma_{K}), \Lambda_{\sigma_P})$. Then, the outer minimum in \eqref{eq:3opt} becomes the minimum of a finite collection of differentiable functions, which enables the \textsf{FWS} algorithm that is tuned for the non-smoothness of the objective function. The assumptions in \cite{wang2021fast} are also satisfied in our problem. As shown in \thmref{thm:lower}, the inner infimum in \eqref{eq:3opt} has a closed form given by $g_P^{\sigma'}(w)$, and by Remark~\ref{rem1}, the objective function of the supremum in \eqref{eq:3opt},  $G_P^{\sigma}(w)$, is a point-wise minimum of continuously differentiable concave functions. 

\subsubsection{Statistics for Stopping}
Recall that \textsf{TaS-FW} employs
\begin{align}
    Z(t) &= t \, G_{\hat{P}(t)}^{\hat{\sigma}(t)}(w(t)) \\
    &= t \, \min_{\sigma \neq \hat{\sigma}(t)} g_{\hat{P}(t)}^{\sigma}(w(t))
\end{align}
as the statistics to decide when to stop. 

As in other existing PE algorithms (e.g., \cite{garivier2016, wang2021fast, prabhu2022, karthik2019}), the statistics $Z(t)$ is related to the GLLR. For any two hypotheses $\sigma_1$ and $\sigma_2$, the GLLR uses the statistics
\begin{align}
    Z_{\sigma_1, \sigma_2}(t) = \log \frac{\max \limits_{P \in \Lambda_{\sigma_1}} \prod_{i = 1}^{K} P_i^{N_i(t)}(\tilde{X}_i^{N_i(t)})}{\max \limits_{P' \in \Lambda_{\sigma_2}} \prod_{i = 1}^{K} {P'_i}^{N_i(t)}(\tilde{X}_i^{N_i(t)})}. \label{eq:Zstat}
\end{align}
The following lemma relates $Z(t)$ with the GLLR statistics.
\begin{lemma}
    It holds that
    \begin{align}
        \max_{\sigma \in \mc{C}} \min_{\sigma' \neq \sigma} Z_{\sigma, \sigma'}(t) = Z(t) - t g_{\hat{P}(t)}^{\hat{\sigma}(t)}(w(t)), \label{eq:Zs1s2}
    \end{align}
    and $\hat{\sigma}(t)$ achieves the maximum on the left-hand side of \eqref{eq:Zs1s2}.
\end{lemma}
\begin{IEEEproof}
    The proof follows from an application of \lemref{lem:GLRG} to the statistics in \eqref{eq:Zstat} and is omitted.
\end{IEEEproof}
The statistics $\max_{\sigma \in \mc{C}} \min_{\sigma' \neq \sigma} Z_{\sigma, \sigma'}(t)$ is used in \cite{garivier2016, agrawalHeavy, agrawal2021, wang2021fast, degenne2019, barrier2022} to decide when to stop. In \cite{prabhu2022, karthik2019}, a modified statistics is used, where on the right-hand side of \eqref{eq:Zstat}, instead of taking the maximization, the likelihood is averaged over the set $\Lambda_{\sigma_2}$ according to a conjugate prior. This modification helps simplify the error probability analysis for vector exponential families.

Given that $\hat{\sigma}(t) = \sigma_P$ for all sufficiently large $t$, $g_{\hat{P}(t)}^{\hat{\sigma}(t)}(w(t))$ approaches 0 for any $w(t) \in \Sigma_{K}$ since $D(\hat{P}_{i_m}(t)\| \hat{P}_{j_m}(t)) \to 0$ for all partition elements $\mc{A}_{m}^{\sigma_P} \in \sigma_P$ and $i_m, j_m \in \mc{A}_{m}^{\sigma_P}$. Then, $ \left(\max_{\sigma \in \mc{C}} \min_{\sigma' \neq \sigma} \frac{Z_{\sigma, \sigma'}(t)}{t}\right) - \frac{Z(t)}{t} \to 0$. The latter also holds for the modified GLLR statistics in \cite{prabhu2022, karthik2019}.

\subsection{Clustering Problems that are not in Our Framework}
Some works that involve clustering with bandit feedback but are not in our framework include \cite{gharat2024, katariya2018, xia2022}.
In \cite{gharat2024}, Gharat et al.\  consider a related PE problem that involves clustering of arms. In their setup, the arms are partitioned into clusters of pre-defined sizes such that for any $j > i$, all arms in cluster $i$ have a larger mean reward than those in cluster $j$, and the goal is to identify a representative arm from each of these clusters; this problem generalizes BAI and the problem of ranking arms~\cite{katariya2018}. In \cite{xia2022}, Xia and Huang study a clustering problem where the decision maker queries whether two chosen arms belong to the same cluster or not and receives a noisy answer to this query; the goal is to identify the hidden partitioning of arms with the least number of queries possible.

\section{Conclusion} \label{sec:conclusion}
In this paper, we introduced a general clustering and distribution matching problem with bandit feedback, where the arm distributions are assumed to have finite alphabets. We developed a framework that enables the analysis of the identification of matching pairs, odd arm, and $N$-ary clusters as a single unified problem. In \thmref{thm:lower}, we derived a non-asymptotic converse bound for any clustering problem in our framework. We developed a computationally-efficient algorithm, which we refer to as \textsf{TaS-FW}, and showed in \thmref{thm:upper} that the average number of arm pulls achieved by this algorithm is asymptotically optimal as the error probability approaches zero. Our bound in \thmref{thm:upper} also includes a novel ``second-order'' term that upper bounds the speed of convergence to the fundamental limit $\frac{\log\frac{1}{\delta}}{T^*(P)}$ as $\delta\to0$. Our simulation results in \secref{sec:experiments} support our theoretical results since the slope of the empirical mean of the number of pulls with respect to $d(\delta \| 1 - \delta)$ on the x-axis is close to the hardness parameter $\frac{1}{T^*(P)}$ in the lower bound. 

\appendices
\renewcommand\thesubsection{\thesection.\Roman{subsection}}
\renewcommand\thesubsectiondis{\thesectiondis.\Roman{subsection}.}

\section{Proofs that Examples 1--3 Satisfy Assumption~\ref{as:finer}} \label{sec:proofAs}
In this section, we prove that Examples 1--3 in \secref{sec:notation} satisfy Assumption~\ref{as:finer}.

\subsection{Example 1}
Since the number of clusters and the size of each cluster in each hypothesis are fixed for all hypotheses ($M$ and 2, respectively), hypothesis $\sigma_1 \in \mc{C}$ dominates hypothesis $\sigma_2 \in \mc{C}$ if and only if $\sigma_1 = \sigma_2$, which proves that this problem satisfies Assumption~\ref{as:finer}.
The scenario where a nominal arm can have an arbitrary number of matches is avoided because that scenario does not satisfy Assumption~\ref{as:finer}.

\subsection{Example 2}
As in Example 1, in this problem, the number of clusters and the sizes of each cluster in each hypothesis are fixed (1 and $K-1$, respectively) for all hypotheses. Therefore, Example 2 satisfies Assumption~\ref{as:finer}.

\subsection{Example 3}
Let $\sigma_1$ be a hypothesis in this problem. Let $|\mc{A}_{M_{\sigma_1}+1}^{\sigma_1}| = k_1$, hence the number of clusters for $\sigma_1$ is $M_{\sigma_1} = N - k_1$. Suppose that $\sigma_1$ dominates some hypothesis $\sigma_2 \neq \sigma_1$, and let $|\mc{A}_{M_{\sigma_2}+1}^{\sigma_2}| = k_2$. This implies that the number of arms that belong to some cluster for $\sigma_2$ is strictly greater than that for $\sigma_1$, which gives $k_2 \leq k_1 - 1$. We consider two cases: $M_{\sigma_2} \leq M_{\sigma_1}$ and $M_{\sigma_2} > M_{\sigma_1}$. In the case of $M_{\sigma_2} \leq M_{\sigma_1}$, the number of groups for $\sigma_2$ is bounded as $M_{\sigma_2} + k_2 \leq M_{\sigma_1} + k_1 - 1 \leq N - 1$. In the other case, let $a = M_{\sigma_2} - M_{\sigma_1} \geq 1$ be the number of newly formed clusters in $\sigma_2$. We notice that each newly formed cluster for $\sigma_2$ must contain at least two arms from the unconstrained group of $\sigma_1$. This argument gives $k_2 \leq k_1 - 2a$. Then, the number of groups for $\sigma_2$ is bounded by $(M_{\sigma_1} + a) + (k_1 - 2a) \leq N - a \leq N - 1$. For both cases, the number of groups for $\sigma_2$ is strictly less than $N$; therefore, $\sigma_2 \notin \mc{C}$, and Example 3 satisfies Assumption~\ref{as:finer}.

\section{Proofs of Lemmas in the Convergence of the \textsf{FWS} Algorithm}
\subsection{Proof of \lemref{lem:gradg}} \label{app:gradg}
Assumption 1 from \cite{wang2021fast} states that for each hypothesis $\sigma \in \mc{C}$, $\Lambda \setminus \Lambda_{\sigma}$ can be written as a union of finitely many convex sets. Under this assumption, Wang et al.\ show in \cite[Prop.~1]{wang2021fast} that the equality in \eqref{eq:gradw} holds. Therefore, it only remains to verify Assumption 1 from \cite{wang2021fast}. In our problem, 
the set $\Lambda_{\sigma}$ is open and convex, and $|\mc{C}|$ finite. Therefore, the set of alternatives to hypothesis $\sigma$, $\Lambda \setminus \Lambda_{\sigma} = \bigcup_{\sigma' \neq \sigma} \Lambda_{\sigma'}$ is a finite union of convex sets, which verifies the assumption. 

From \eqref{eq:gradw}, we deduce that $\lVert \nabla_w g_P^{\sigma'}(w) \rVert_{\infty} \leq \max_{m \in [M]} \max_{i_m \in \mc{A}_{m}^{\sigma'}} D(P_{i_m} \| W_m) \leq L$, and the $L$-Lipschitzness of $g_P^{\sigma'}(w)$ in $w$ follows from the mean value theorem. 

Then, for all $w, \tilde{w} \in \Sigma_K$, 
\begin{align}
    G_P^{\sigma}(\tilde{w}) &= \min_{\sigma' \neq \sigma} g_P^{\sigma'}(\tilde{w}) \geq \min_{\sigma' \neq \sigma} \left( g_P^{\sigma'}(w) - L \norm{\tilde{w} - w}_{\infty} \right) \label{eq:GP1} \\
    &= \left( \min_{\sigma' \neq \sigma} g_P^{\sigma'}(w) \right) - L \norm{\tilde{w} - w}_{\infty} \\
    &= G_P^{\sigma}(w) - L \norm{\tilde{w} - w}_{\infty}, \label{eq:GP3}
\end{align}
which proves the $L$-Lipschitzness of $G_P^{\sigma}(w)$ in $w$. 

\subsection{Proof of \lemref{lem:gradLip}} \label{app:gradLip}
The proof is inspired by \cite[Appendix C.1]{wang2021fast}, which is for BAI.
    We first prove that $\lVert \nabla^2_{w w} g_P^{\sigma'}(w) \rVert_{\infty} \leq \frac{D}{\gamma}$.
    By \eqref{eq:gradw} from \lemref{lem:gradg}, for every $i \in [K]$, $(\nabla^2_{w w} g_P^{\sigma'}(w))_{i, j}$ is non-zero only for the values $j \in [K]$ where $i, j \in \mc{A}_m^{\sigma'}$ for some $m \in [M]$. 

    Consider $(i_m, j_m) \in [K]^2$ such that $i_m$ and $j_m$ belong to the same cluster $m$ for the hypothesis $\sigma'$, i.e., $i_m, j_m \in \mc{A}_m^{\sigma'}$. In \eqref{eq:gradw}, differentiating the term corresponding to $e_{i_m}$ with respect to $w_{j_m}$ gives the Hessian
    \begin{align}
        &\nabla^2_{w w} g_P^{\sigma'}(w)_{i_m, j_m} \notag \\
        &= \sum_{a \in \mc{X}} \frac{\partial}{\partial W_m(a)} D(P_{i_m} \| W_m) \cdot \frac{\partial}{\partial w_{j_m}} W_m(a) \\
        &= - \sum_{a \in \mc{X}} \frac{P_{i_m}(a)}{W_m(a)} \cdot \frac{\sum_{j \in \mc{A}_m^{\sigma'} \setminus {j_m}} w_{j} (P_j(a) - P_{j_m}(a))}{(\sum_{i \in \mc{A}_m^{\sigma'}} w_{i})^2}.
    \end{align}
    For $\gamma > 0$, we have
    \begin{align}
        \max_{w_a, w_b \geq \gamma} \frac{w_b}{(w_a + w_b)^2} = \frac{1}{4 \gamma}. \label{eq:w1w2gam}
    \end{align}
    Since $p_{\min} \leq P_i(x) \leq 1 - p_{\min}$ for all arms $i \in [K]$, we have $W_m(a) \geq p_{\min}$ for all $m$. Using \eqref{eq:w1w2gam} with $w_a \gets w_{j_m}$ and $w_b \gets \sum_{j \in \mc{A}_{m}^{\sigma'} \setminus j_m} w_j$, we reach
    \begin{align}
        \left \lvert \nabla^2_{w w} g_P^{\sigma'}(w)_{i_m, j_m} \right \rvert \leq |\mc{X}| \frac{1 - p_{\min}}{4 p_{\min} \gamma}.
    \end{align}
    and
    \begin{align}
        \norm{\nabla^2_{w w} g_P^{\sigma'}(w)}_\infty \leq \frac{D}{\gamma}
    \end{align}
    where the latter follows since every row of the Hessian has at most $\max_{\sigma' \in \mc{C}} \max_{m \in [M]}|\mc{A}_m^{\sigma'}|$ non-zero values. The gradient $\nabla_w g_P^{\sigma'}(w)$ is $\frac{D}{\gamma}$-Lipschitz as claimed.

\subsection{Proof of \lemref{lem:LipinP}} \label{app:LipinP}
    The proof follows steps similar to those in \cite[Lemma~14]{wang2021fast}. In particular, we use the closed-form expression for the partial derivative $\frac{\partial g_P^{\sigma}(w)}{\partial P_i(a)}$.
    
    Fix a hypothesis $\sigma \in \mc{C}$. For $P = Q$, the claim trivially holds. Assume that $P \neq Q$. Define the function $f \colon [0, 1] \to \mathbb{R}$ as
    \begin{align}
        f(t) = g_{tP + (1-t)Q}^{\sigma}(w).
    \end{align}
    Let $\frac{\partial g_{P}^{\sigma}(w)}{\partial P}(w', P')$ be the partial derivative of $g_{P}^{\sigma}(w)$ with respect to $P$, evaluated at $w = w'$ and $P = P'$.
    
    From the mean value theorem, there exists some $t \in (0, 1)$ such that
    \begin{align}
        f'(t) = g^{\sigma}_P(w) - g^{\sigma}_Q(w). 
    \end{align}
    Hence,
    \begin{align}
        g^{\sigma}_P(w) - g^{\sigma}_Q(w) = \left\langle 
        \frac{\partial g_P^{\sigma}(w)}{\partial P}(w, \tilde{Q}), P - Q \right\rangle.  \label{eq:gabs}
    \end{align}
    Let $\tilde{Q} = t P + (1-t) Q$. We compute for $i \in [K]$ and $a \in \mc{X}$
    \begin{align}
        \frac{\partial g_P^{\sigma}(w)}{\partial P_i(a)}(w, \tilde{Q}) = w_i \left( \log \frac{\tilde{Q}_i(a)}{\tilde{W}_i(a)} + 1 \right), \label{eq:partialcomp}
    \end{align}
    where
    \begin{align}
        \tilde{W}_i(a) = \frac{\sum_{i_m \in \mc{A}_m^{\sigma}} w_{i_m} \tilde{Q}_{i_m}(a)}{\sum_{i_m \in \mc{A}_m^{\sigma}} w_{i_m}} 
    \end{align}
    and $m \in [M]$ is the unique cluster index such that $i \in \mc{A}_m^{\sigma}$.

    Combining \eqref{eq:gabs} and \eqref{eq:partialcomp}, we get
    \begin{align}
         &|g^{\sigma}_P(w) - g^{\sigma}_Q(w)| \notag \\
         &= \left \lvert \sum_{i = 1}^K w_i \sum_{a \in \mc{X}}  \left(\log \frac{\tilde{Q}_i(a)}{\tilde{W}_i(a)} + 1 \right) (P_i(a) - Q_i(a)) \right \rvert\\ 
         &\leq \sum_{i = 1}^K w_i |\mc{X}| \log \frac{1 - p_{\min} + \epsilon}{p_{\min} - \epsilon} \epsilon  \label{eq:stepb}  \\
         &\leq  |\mc{X}| \log \frac{2 - p_{\min}}{p_{\min}} \epsilon, \label{eq:pmineps}
    \end{align} 
    where in \eqref{eq:stepb}, we use the fact that $\sum_{a \in \mc{X}} P_i(a) - Q_i(a) = 0$ for all $i \in [K]$, $|P_i(a) - Q_i(a)| \leq \epsilon$ for all $i \in [K]$ and $a \in \mc{X}$, and that 
    \begin{align}
        p_{\min} - \epsilon &\leq \tilde{Q}_i(a) \leq 1 - p_{\min} + \epsilon \\
        p_{\min} - \epsilon &\leq \tilde{W}_i(a) \leq 1 - p_{\min} + \epsilon.
    \end{align}
    In \eqref{eq:pmineps}, we use the fact that $\epsilon \leq \frac{p_{\min}}{2}$.
    
    We note that
    \begin{align}
        G_Q^{\sigma}(w) = \min_{\sigma' \neq \sigma} g_Q^{\sigma'}(w),
    \end{align}
    where $g_Q^{\sigma'}(w) \in [g_P^{\sigma'}(w) - E \epsilon, g_P^{\sigma'}(w) + E \epsilon]$ for all $\sigma' \neq \sigma$. Hence, \eqref{eq:GPLipP} follows.

    \subsection{Proof of \lemref{lem:maxminz}} \label{app:maxminz}
    The only difference between the proofs of \lemref{lem:maxminz} and \cite[Th.~6]{wang2021fast} is that due to the forced exploration indices $\mc{I}_{\mr{f}}$ defined in \eqref{eq:Ifdef}, $\tilde{x}_i(t-1) = \Omega(t^{-1/2} \log t)$, while in \cite{wang2021fast}, $\tilde{x}_i(t-1) = \Omega(t^{-1/2})$ holds. We provide the proof for completeness.
    
    From \eqref{eq:If} and \eqref{eq:ztilde}--\eqref{eq:xtildedef}, we have
    \begin{align}
        \tilde{x}_i(t-1) \geq \frac{({t-1})^{-1/2} \log(t-1)}{K}
    \end{align}
    for all $i \in [K]$.
    For $t \geq 3$, we have
    \begin{align}
        \tilde{x}_i(t-1) \geq \frac{{t}^{-1/2} \log t}{2K}.
    \end{align}
    Hence, for $t \geq K + 1 \geq 3$, $\tilde{x}(t-1) \in \Sigma_K^{\frac{{t}^{-1/2} \log t}{2K}}$.
    
    Putting $\alpha \gets \frac{1}{t}$, $x \gets \tilde{x}(t-1)$, $y \gets \tilde{x}(t)$, $z \gets \tilde{z}(t)$, $r \gets r_t$, and $\gamma \gets \frac{{t}^{-1/2} \log t}{2K}$ in \lemref{lem:GPoracle}, we get
    \begin{align}
        &G_P(\tilde{x}(t)) - G_P(\tilde{x}(t-1)) \notag \\
        &\geq \frac{1}{t} \min_{h \in H_{G_P}(\tilde{x}(t-1), r_t)} \langle \tilde{z}(t) - \tilde{x}(t-1), h \rangle - \frac{16 D K}{t^{3/2} \log t} \\
        &\geq \frac{1}{t} \left( \max \limits_{z \in \Sigma_{K}} \min \limits_{h \in H_{G_P}(\tilde{x}(t-1), r_t)} \langle z - \tilde{x}(t-1), h \rangle - \epsilon_t \right) \notag \\
        &\quad - \frac{16 D K}{t^{3/2} \log t}\label{eq:2oracle} \\
        &\geq \frac{1}{t} \left( \max \limits_{z \in \Sigma_{K}} \min \limits_{h \in \partial_{r_t} {G_P}(\tilde{x}(t-1))} \langle z - \tilde{x}(t-1), h \rangle - \epsilon_t \right) \notag \\
        &\quad - \frac{16 D K}{t^{3/2} \log t}\label{eq:Partialoracle} \\
        &\geq \frac{1}{t} \left( \Delta_{t-1} - r_t - \epsilon_t \right) -  \frac{16 D K}{t^{3/2} \log t}, \label{eq:3oracle}
    \end{align}
    where $\partial_{r}(\cdot)$ is the $r$-subdifferential defined in \eqref{eq:rsubdiff}. Inequality
    \eqref{eq:2oracle} follows by the assumption on $\tilde{z}(t)$ and $\tilde{x}(t)$. Inequality \eqref{eq:Partialoracle} follows since $H_{G_P}(\tilde{x}(t-1), r_t) \subseteq \partial_{r_t} {G_P}(\tilde{x}(t-1))$ \cite[Lemma~10]{wang2021fast}, and  \eqref{eq:3oracle} is stated in \cite[Lemma~11]{wang2021fast} for functions that are minimum of concave functions. The proof is completed by subtracting $G_P(w^*)$ from both sides of \eqref{eq:3oracle}.

    \subsection{Proof of \lemref{lem:T2gap}} \label{app:T2gap}
    By the Lipschitzness of $G_P(\cdot)$ shown in \lemref{lem:gradg}, we have
    \begin{align}
        T_1 \Delta_{T_1} \leq  T_1 L. \label{eq:T1L}
    \end{align}
    Multiplying both sides of \eqref{eq:Deltat} by $t$, we get
    \begin{align}
          t \Delta_t \leq (t-1) \Delta_{t-1} + r_t + \epsilon_t + \frac{16 D K}{t^{1/2} \log t} \label{eq:tDeltat}
    \end{align}
    for $t \in \{T_1 + 1, \dots, T_2\} \cap \mc{I}_{\mr{f}}^{\mr{c}}$. For $t \in \{T_1 + 1, \dots, T_2\} \cap \mc{I}_{\mr{f}}$, we have
    \begin{align}
        \tilde{x}(t) = \frac{1}{t} \left(\frac{1}{K}, \dots, \frac{1}{K}\right) + \frac{t-1}{t} \tilde{x}(t-1),
    \end{align}
    which gives $\lVert\tilde{x}(t) - \tilde{x}(t-1) \rVert_{\infty} \leq \frac{1}{t}$. We apply \lemref{lem:gradg} to get
    \begin{align}
        \Delta_t \leq \Delta_{t-1} + \frac{L}{t} \label{eq:DtIf}
    \end{align}
    for $t \in \mc{I}_{\mr{f}}$.
    Since $\Delta_{t-1} \leq L$, we get from \eqref{eq:DtIf}
    \begin{align}
        t \Delta_t &\leq (t-1) \Delta_{t-1} + 2 L \\
        &\leq (t-1) \Delta_{t-1} + 2 L + 
        r_t + \epsilon_t + \frac{16 D K}{t^{1/2} \log t} \label{eq:tDtIf}
    \end{align}
    for $t \in \{T_1 + 1, \dots, T_2\} \cap \mc{I}_{\mr{f}}$.
    Summing the inequalities in \eqref{eq:T1L}, \eqref{eq:tDeltat} and \eqref{eq:tDtIf}, we get
    \begin{align}
        T_2 \Delta_{T_2} &\leq T_1 L + 2 L |\mc{I}_{\mr{f}} \cap [T_2]| + \sum_{t = T_1 + 1}^{T_2} \left(r_t + \epsilon_t +  \frac{16DK}{t^{1/2} \log t} \right)\\
        &\leq T_1 L + 2L (T_2^{1/2} \log T_2 + 2) \notag \\
        &\quad + \sum_{t = 1}^{T_2} \left(r_t + \epsilon_t\right) + \int_{0}^{T_2} \frac{16DK}{t^{1/2}}  \mr{d}t \label{eq:stepIfint}\\
        &\leq T_1 L + 2L T_2^{1/2} \log T_2 + \sum_{t = 1}^{T_2} \left(r_t + \epsilon_t\right) \notag \\
        &\quad + 32 D K T_2^{1/2} + 4L,
    \end{align}
   where \eqref{eq:stepIfint} follows from \eqref{eq:If} and the fact that $\log t \geq 1$ for $t \geq T_1 + 1$.

   \section{Proof of \lemref{lem:concent}} \label{app:concent}
   The proof generalizes \cite[Lemma~5]{haghifam} to a scenario with $M \geq 1$ clusters with arbitrary sizes.

Define 
\begin{align}
    \mc{R} &= \bigg\{V = (V_1, \dots, V_K) \in \prod_{i = 1}^K \mc{P}_{n_i}(\mc{X}) \colon \notag \\
    &\quad N \sum_{m = 1}^{M} G(V_{\mc{A}_m}, w_{\mc{A}_m}) \geq \beta\bigg\}
\end{align} 
and
\begin{align}
    W_m(w, V) = \frac{\sum_{i_m \in \mc{A}_m} w_{i_m} V_{i_m}}{\sum_{i_m \in \mc{A}_m} w_{i_m}}.
\end{align}
Using standard bounds based on the method of types, we have
\begin{align}
    &\Prob{N \sum_{m = 1}^{M} G(\hat{P}_{\mc{A}_m}, w_{\mc{A}_m}) \geq \beta} \notag \\
    &\leq \sum_{V \in \mc{R}} \exp\left\{ - \sum_{i = 1}^K n_i D(V_i \| P_i) \right\} \label{eq:mot1}\\
    &= \sum_{V \in \mc{R}} \exp\bigg \{- N \sum_{m = 1}^M G(V_{\mc{A}_m}, w_{\mc{A}}) \notag \\
    &\quad \quad - \sum_{m = 1}^M \sum_{i_m \in \mc{A}_{m}} n_{i_m} D( W_m(w, V) \| Q_m) \bigg\} \label{eq:KLidentity} \\
    &\leq \exp\{-\beta\} \prod_{m = 1}^M \left(\sum_{i_m \in \mc{A}_{m}} n_{i_m} + 1\right)^{|\mc{X}|} \notag \\
    &\quad \cdot  \sum_{V \in \mc{R}} \Prob{ \bigcap_{m = 1}^M (X_{i_m}^{n_{i_m}} \colon i_m \in \mc{A}_m) \in \mc{T}_{W_m(w, V)}^{\sum_{i_m \in \mc{A}_m} n_{i_m}}  } \label{eq:mot2}\\
    &\leq \exp\{-\beta\} (N + 1)^{M |\mc{X}|},
\end{align}
where \eqref{eq:mot1} and \eqref{eq:mot2} follow from the upper and lower bounds in \cite[Th.~11.1.4]{cover}, respectively. The equality \eqref{eq:KLidentity} follows from the identity 
\begin{align}
    &\sum_{i_m \in \mc{A}_m} w_{i_m} D(V_{i_m} \| P_i) \notag \\
    &= \sum_{i_m} w_{i_m} (D(V_{i_m} \| W_m(w, V)) + D(W_m(w, V) \| Q_m)). 
\end{align}

\section{Proof of \eqref{eq:GPresult2}} \label{app:GPresult}
In the following, we show that on the event $\mc{E}(T)$ in \eqref{eq:ETdef}, it holds that
\begin{align}
    G_P^{\sigma_P}(w(t)) - E \epsilon_t \geq T^*(P) - \psi(t). \label{eq:proofGP}
\end{align}
By \eqref{eq:E1t} and \eqref{eq:ETdef}, on the event $\mc{E}(T)$, the condition \eqref{eq:highprobevent} in Corollary~\ref{cor:GP} is satisfied for $t \in [h_1(T), T]$. Hence, applying Corollary~\ref{cor:GP} to $G_P^{\sigma_P}(w(t))$ with $T_2$ replaced by $t$,  $T_1$ replaced by $h_1(T) \leq T^{b_1} + 1$, and $\epsilon_t$ replaced by $2 E \epsilon_t$ yields
\begin{align}
    G_P^{\sigma_P}(w(t)) &\geq T^*(P) - \bigg[ \frac{T^{b_1} + 1}{t} L + 2L t^{-1/2} \log t \notag \\
    &\quad + \frac{1}{t}\sum_{s = 1}^{t} \left(r_s + 2 E \epsilon_s \right) \notag \\
        &\quad + 32 D K t^{-1/2} + \frac{L(K + 3)}{t} \bigg]. \label{eq:GPboundcor}
\end{align}
Since $t \geq h_2(T) \geq T^{b_2}$, 
\begin{align}
    \frac{T^{b_1}}{t} \leq t^{-\frac{b_2 - b_1}{b_2}}. \label{eq:Tb1}
\end{align}
We bound the summation in \eqref{eq:GPboundcor} as
\begin{align}
     \frac{1}{t}\sum_{s = 1}^{t} \left(r_s + 2 E \epsilon_s \right) &\leq \frac{1}{t} \int_{0}^t (s^{-b_0} + 2 E \epsilon_s) \mr{d}s \\
     &\leq \frac{1}{1 - b_0} t^{-b_0} + \frac{2 E \sqrt{\log t}}{t} \int_{0}^t F s^{-1/4} \mr{d}{s} \\
     &= \frac{1}{1- b_0} t^{-b_0} + \frac{8 E F}{3} t^{-1/4} \sqrt{\log t}, \label{eq:sumrtet}
\end{align}
where we use \eqref{eq:rtb0} and \eqref{eq:et}. Combining \eqref{eq:et}, \eqref{eq:GPboundcor}, \eqref{eq:Tb1}, and \eqref{eq:sumrtet}, we get \eqref{eq:proofGP}.

\section*{Acknowledgment}

We thank the anonymous reviewers for their constructive suggestions, which resulted in the addition of Sections~\ref{sec:Gaussian} and~\ref{sec:comp}.

J.~Scarlett was supported by the Singapore National Research Foundation under its AI Visiting Professorship programme.

\bibliographystyle{IEEEtran}
\bibliography{mac} 

\begin{IEEEbiographynophoto}{Recep Can Yavas}
(Member, IEEE) received the B.S. degree (Hons.) in electrical engineering from Bilkent University, Ankara, Turkey, in 2016. He received the M.S. and Ph.D. degrees in electrical engineering from California Institute of Technology (Caltech) in 2017 and 2023, respectively. He was a research fellow at CNRS@CREATE, Singapore, between 2022 and 2024. In October 2024, he joined the Department of Computer Science at National University of Singapore as a research fellow. 
His research interests include information theory, communications, and multi-armed bandits.
\end{IEEEbiographynophoto}

\begin{IEEEbiographynophoto}{Yuqi Huang} is currently a senior undergraduate student in business analytics at National University of Singapore.
His research interests include machine learning and multi-armed bandits.
\end{IEEEbiographynophoto}

\begin{IEEEbiographynophoto}{Vincent Y. F. Tan} (Senior Member, IEEE) was born in Singapore, in 1981.
He received the B.A. and M.Eng. degrees in electrical and information
science from Cambridge University in 2005 and the Ph.D. degree in electrical
engineering and computer science (EECS) from Massachusetts Institute of
Technology (MIT) in 2011. He is currently a Professor with the Department of Mathematics and the
Department of Electrical and Computer Engineering (ECE), National University of Singapore (NUS). His research interests include information theory,
machine learning, and statistical signal processing. 

He is an elected member
of the IEEE Information Theory Society Board of Governors. He was an
IEEE Information Theory Society Distinguished Lecturer from 2018 to 2019.
He received the MIT EECS Jin-Au Kong Outstanding Doctoral Thesis Prize
in 2011, the NUS Young Investigator Award in 2014, Singapore National
Research Foundation (NRF) Fellowship (Class of 2018), and the NUS Young
Researcher Award in 2019. He is also serving as a Senior Area Editor for
{\em IEEE Transactions on Signal Processing} and an Associate Editor in
Machine Learning and Statistics for {\em  IEEE Transactions on  Information Theory}. He also regularly serves as the Area Chair for Prominent
Machine Learning Conferences, such as the International Conference on
Learning Representations (ICLR) and the Conference on Neural Information
Processing Systems (NeurIPS).
\end{IEEEbiographynophoto}

     \begin{IEEEbiographynophoto}{Jonathan Scarlett}
        (Member, IEEE) received 
        the B.Eng.~degree in electrical engineering and the B.Sci.~degree in 
        computer science from the University of Melbourne, Australia. 
        From October 2011 to August 2014, he
        was a Ph.D. student in the Signal Processing and Communications Group
        at the University of Cambridge, United Kingdom. From September 2014 to
        September 2017, he was post-doctoral researcher with the Laboratory for
        Information and Inference Systems at the \'Ecole Polytechnique F\'ed\'erale
        de Lausanne (EPFL), Switzerland. Since January 2018, he has been with the Department of Computer Science and Department of Mathematics at the
        National University of Singapore, where he is currently an Associate Professor. His research interests are in
        the areas of information theory, machine learning, signal processing, and
        high-dimensional statistics. He received the Singapore National Research Foundation (NRF)
        fellowship and the NUS Presidential Young Professorship award, and he is currently serving as an Associate Editor for \emph{IEEE Transactions on Information Theory}.
    \end{IEEEbiographynophoto}
    
\end{document}